\documentclass{article}

\usepackage[svgnames]{xcolor} 
\usepackage{graphicx}
\usepackage[margin= 0.8in]{geometry}
\usepackage{changepage}

\newcommand*{\titleAT}{\begingroup 
\parindent0pt
\newlength{\drop} 
\drop=0.05\textheight 

\sffamily

\includegraphics[scale=1.5]{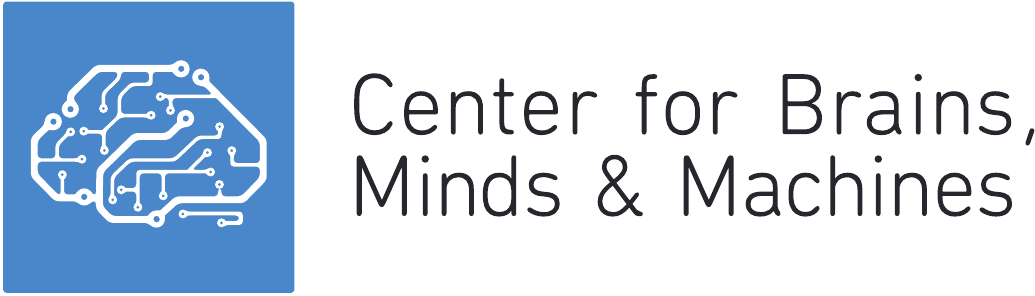}

\textcolor{CornflowerBlue}{\rule{\textwidth}{3 pt}}\par 
\vspace{2pt}\vspace{-\baselineskip} 
\rule{\textwidth}{0.4pt}\par 

\vspace{\drop} 
\textbf{\large{CBMM Memo No. \memonumber}}\hfill\textbf{\large{\memodate}}

\vspace{\drop}
\begin{center}
\textbf{\huge{\memotitle}}\\
\vspace{0.4\drop}
\textbf{\Large{by}}\\
\vspace{0.4\drop}
\textbf{\large{\memoauthors}}
\end{center}
\vspace{\drop}
\begin{adjustwidth}{2.5em}{0pt}
\textbf{\large{\noindent Abstract}:} {\memoabstract}
\end{adjustwidth}

\vfill
\textcolor{CornflowerBlue}{\rule{\textwidth}{3 pt}}\par 
\vspace{2pt}\vspace{-\baselineskip} 
\rule{\textwidth}{0.4pt}\par

\vspace{3pt}
\begin{minipage}{.15\linewidth}
\includegraphics[scale=0.1]{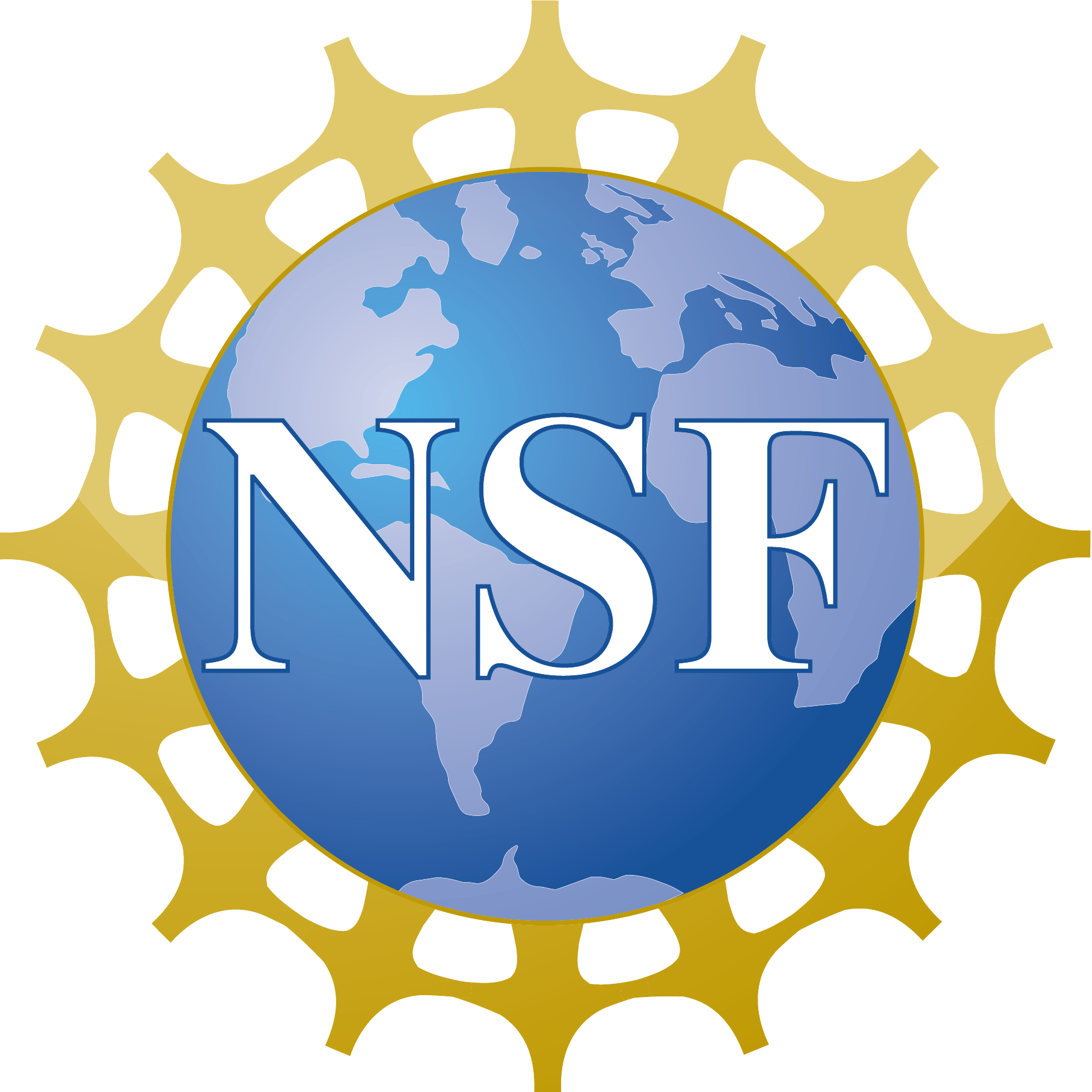}
\end{minipage}
\begin{minipage}{.84\linewidth}
\large{This work was supported by the Center for Brains, Minds and Machines (CBMM), funded by NSF STC award  CCF-1231216.}
\end{minipage}
\thispagestyle{empty}
\endgroup}

\RequirePackage[OT1]{fontenc}
\RequirePackage{amsthm,amsmath,amssymb}
\RequirePackage[square,authoryear,sort]{natbib}
\RequirePackage[authoryear]{natbib}
\RequirePackage[colorlinks,citecolor=blue,urlcolor=blue]{hyperref}

\usepackage{macros}
\usepackage[mathcal, mathscr]{eucal}
\usepackage{colortbl}
\usepackage[size=small]{caption}
\usepackage{subcaption}
\usepackage{graphicx}
\DeclareGraphicsExtensions{.pdf,.png,.jpg,.eps}
\usepackage{algorithm}
\usepackage{algorithmic}
\usepackage[authoryear,sort]{natbib}
\setcitestyle{authoryear,square}

\bibpunct[; ]{(}{)}{;}{a}{,}{;}

\usepackage{booktabs}
  
\begin{document}

\def\memonumber{ 027} 
\def\memodate{\today} 
\def\memotitle{A Nonparametric Bayesian Approach to Uncovering Rat Hippocampal Population Codes During Spatial Navigation} 
\def\memoauthors{Scott W. Linderman, Matthew J. Johnson, Matthew A. Wilson, and Zhe Chen}
\def\memoabstract{
Rodent hippocampal population codes represent important spatial  information about the environment during navigation. 
Several computational methods have been developed to uncover the neural representation of spatial topology embedded in rodent hippocampal 
ensemble spike activity. Here we extend our previous work and propose a nonparametric Bayesian approach to infer rat hippocampal population codes during spatial navigation.
To tackle the model selection problem, we leverage a nonparametric Bayesian model. 
Specifically, to analyze  rat hippocampal 
ensemble spiking activity, we apply a hierarchical Dirichlet process-hidden Markov model (HDP-HMM) using two Bayesian inference methods, one based on Markov chain Monte Carlo (MCMC) and the other based on variational Bayes (VB). We demonstrate the effectiveness of our Bayesian approaches on recordings from a freely-behaving rat navigating in an open field environment. We find that MCMC-based inference with Hamiltonian Monte Carlo (HMC) hyperparameter sampling is flexible and efficient, and outperforms VB and MCMC approaches with hyperparameters set by empirical Bayes.
}

\titleAT 

\section{Introduction}
A fundamental goal in neuroscience is to understand how populations of  neurons represent and transmit information about the external world. 
The hippocampus is known to encode information relevant to spatial navigation and episodic memory. 
Spatial representation of the environment is pivotal for navigation in rodents \citep{OKeefe78}.
One type of spatial representation is a topological map, which contains only relative ordering or connectivity information
between spatial locations and is invariant to orientation or deformation. A relevant question of interest is: how can neurons 
downstream of the hippocampus  infer representations of space from 
hippocampal spike  activity without {\em a priori} place field information (namely, without the measurement of spatial correlates)?  
Several reports have been dedicated to the mathematical analysis of this problem \citep{Curto08,Dabaghian12}; however, a data-driven approach 
for analyzing ensemble hippocampal spike data remains missing.  This paper 
employs probabilistic modeling and inference methods to uncover the spatial representation (or topological map) based on the ensemble spike activity.

Bayesian statistical modeling is a consistent and principled  framework for dealing with uncertainties 
about the observed data \citep{Scott02}.  
The goal of Bayesian inference is to incorporate prior knowledge and constraints
of the problem and to infer the posterior distribution of  unobserved variables of interest \citep{Gelman13}. In recent years, cutting-edge Bayesian methods have become increasingly popular
for data analyses in neuroscience, medicine and biology \citep{Mishchenko11a,Chen11,Chen13,Davidson09,Kloosterman14,Yau11}. Specifically, thanks to ever-growing computing power, Markov chain Monte Carlo (MCMC) methods have been widely used in Bayesian inference. 

In our previous work \citep{Chen12a,Chen14}, we have developed a {\em parametric Bayesian} approach to uncover the neural 
representation of spatial topology embedded in rodent hippocampal population codes during spatial navigation. 
Here we extend the preceding work and consider a {\em nonparametric Bayesian} 
approach. 
The nonparametric Bayesian method brings additional flexibility to the probabilistic model, which allows us 
to model the complex structure of neural data \citep{Teh10,Wood08,Yu09,Shalchyan14}.  
Specifically, we leverage the so-called a hierarchical Dirichlet process-HMM (HDP-HMM) \citep{Teh06}, which extends the finite-state hidden Markov model (HMM) with a nonparametric, HDP prior and derive corresponding Bayesian inference algorithm. 
We consider both deterministic and stochastic approaches for fully Bayesian inference. 
Based on deterministic approximation, we extend the work of \citep{Johnson14,Chen12a} and use a variational Bayes (VB) method for approximate Bayesian inference.  For MCMC, we adapt a Gibbs sampling method \citep{Johnson14b, Teh06}, and integrate it with a Hamiltonian  Monte Carlo (HMC) method for hyperparameter inference\citep{Neal10}. 
To the best of our knowledge, the application of the HDP-HMM to hippocampal ensemble neuronal spike trains and the HMC hyperparameter inference algorithm is novel.

We test the statistical model and inference methods with  both simulation data and experimental data. The latter consists of a recording of rat dorsal hippocampal ensemble spike activity during  open field navigation. Using a decoding analysis and predictive likelihood, we verify and  compare the performance of the proposed Bayesian inference algorithms. We also discuss the results of model selection related to the sample size and the choice of concentration parameter or hyperparameters. Our methods provide an extended tool to analyze rodent hippocampal population codes, which may further empower us to explore important neuroscience questions about neural representation, learning and memory.

\section{Methods: Modeling and Inference}

\subsection{Basic Probabilistic Model}

In our previous work  \citep{Chen12a}, we  used a finite $m$-state HMM to characterize the population spiking activity from a population of $C$ hippocampal place cells. 
It was assumed that first, the animal's spatial location during locomotion, modeled as a latent state process, followed a first-order discrete-state Markov chain $\mathcal{S}=S_{1:T}\equiv \{S_t\}\in\{1,\dots, m\}$, and second, the spike counts of individual place cells at time $t$, conditional on the hidden state $S_t$, followed a Poisson  probability with their respective tuning curve functions ${\matr\Lambda=\{\vect\lambda_c\}=\{\{\lambda_{c,i}\}\}}$. Essentially, we employed a Markov-driven population Poisson firing model with the following probabilistic models 
\begin{align}
p(\by_{1:T}, S_{1:T} \given \bpi, \bP, \bLambda) &= p(\mathcal{S}_1\given \bpi) \prod_{t=2}^T p(S_t \given S_{t-1}, \bP) \prod_{t=1}^T p(\by_t \given S_t, \bLambda),\\
\nonumber p(S_1 \given \bpi) &= \texttt{Multinomial}(S_1 \given \bpi), \\
\nonumber p(S_t \given S_{t-1}, \bP) &= \texttt{Multinomial}(S_t \given \bP_{S_{t-1},:}), \\
\nonumber p(\by_t \given S_t, \bLambda) &= \prod_{c=1}^C \texttt{Poisson}(y_{c,t} \given \lambda_{c,S_t}).
\end{align}
where  $\matr P=\{P_{ij}\}$ denotes an $m$-by-$m$ state transition probability matrix, with $P_{ij}$ representing the transition probability from state $i$ to $j$ (since $\sum_{k=1}^m P_{ik}=1$, each row of matrix $\matr P$ specifies a multinomial likelihood); $y_{c,t}$ denotes the number of spike counts from the $c$-th cell within the $t$-th temporal bin (for notation simplicity, from now on we assume $\Delta=1$ such  that the rate is defined in the unit bin size of 250 ms) and ${\by}_{1:T}=\{y_{c,t}\}_{C\times T}$ denotes time series of $C$-dimensional  population response vector; and $\texttt{Poisson}(y_{c,t}|\lambda_{c,i})$ defines a Poisson distribution with the rate parameter $\lambda_{c,i}$ when $S_t=i$. Finally,  $\log p({\by}_{1:T}|\mathcal{S},\vect\theta)$ defines the observed data log likelihood given the hidden state sequence $\mathcal{S}$ and all parameters $\vect\theta=\{\vect\pi, \matr P, \matr\Lambda\}$ (where $\vect\pi=\{\pi_i\}$ denotes a probability vector for the initial state $S_1$).   

The hidden variables $\mathcal{S}=\{S_{1:T}\}$ are treated as the missing data, ${\by}_{1:T}$ as the observed (incomplete) data, and  their combination $\{S_{1:T},{\by}_{1:T}\}$ as the complete data.

A Bayesian version of this model introduces prior distributions over the parameters. We use the following conjugate prior distributions,
\begin{align}
\bpi &\sim \texttt{Dir}(\alpha_0\bone),\\
\nonumber \bP_{i,:} &\sim \texttt{Dir}(\alpha_0\bone), \\
\nonumber \lambda_{c,i} &\sim \texttt{Gamma}(a_c^0,b_c^0).
\end{align}
where $\texttt{Dir}$ denotes the Dirichlet prior distribution, and $\texttt{Gamma}(a_c^0,b_c^0)$ denotes the gamma prior distribution with 
shape parameter $a_c^0$ and scale parameter $b_c^0$.

\subsection{HDP-HMM}

Model selection is an important issue for statistical modeling and data analysis. 
We have previously proposed a {\em Bayesian deviance information criterion} to select the model size $m$  of HMM \citep{Chen12a,Chen14}. Here we extend the finite-state HMM to an HDP-HMM, a nonparametric Bayesian extension of the HMM that allows for a potentially infinite number of hidden states \citep{Beal02}. Namely,  the HDP-HMM treats the priors via a stochastic process. Instead of imposing a Dirichlet prior distribution on the rows of the finite state transition matrix $\matr P$, we use a HDP that allows for a countably infinite number of states. 

Specifically, we sample a distribution over latent states,~$G_0$, from a Dirichlet process (DP) prior,~${G_0\sim\texttt{DP}(\gamma,H)}$, where~$\gamma$ is the concentration parameter and~$H$ is the base measure. One may sample from the DP via the ``stick-breaking construction.'' First we sample the stick-breaking weights,~$\bbeta$, 
\begin{eqnarray}                                   
\tilde{\beta}_{i}\sim \texttt{Beta}(1,\gamma), \;\;\; \beta_{i}=\tilde{\beta}_{i}, \;\;\; \beta_{i}=\tilde{\beta}_{i}\prod_{k=1}^{j-1}\Big(1-\tilde{\beta}_{i}\Big)
\label{stick1}
\end{eqnarray}
where $\sum_{j=1}^\infty \beta_{i}=1$, and $\texttt{Beta}(a,b)$ defines a beta distribution with two positive shape parameters $a$ and $b$. 

The stick-breaking construction of (\ref{stick1}) is generally denoted as~${\vect\bbeta \sim \texttt{GEM}(\gamma)}$, after Griffiths, Engen, and McCloskey \citep{Ewens90}. 
The name ``stick-breaking'' comes from the interpretation of $\beta_i$ as the length of the piece of a unit-length stick assigned to the $i$-th value.
After the first $i-1$ values having their portions assigned, the length of the remainder of the stick is broken according  to a sample $\tilde{\pi}_i$
from a beta distribution, and $\tilde{\beta}_i$ indicates the portion of the remainder to be assigned to the $i$-th value. Therefore, the stick-breaking process $\texttt{GEM}(\gamma)$ also defines a DP---the smaller $\gamma$, the less (in a statistical sense) 
of the stick will be left for subsequent values. 

After sampling~$\bbeta$, we next sample the latent state variables, in this case~${\blambda_c}$, from the base measure~$H$. Our draw from the~${\mathrm{DP}(\gamma,H)}$ prior is then given by
\begin{align}
G_0=\sum_{j=1}^\infty \beta_j \delta_{\blambda_{c}^{(j)}}.
\end{align}
Importantly, this distribution is discrete with probability one \citep{Teh06}.

Given a countably infinite set of shared states, we may then sample the rows of the transition matrix,~${\matr P_{i,:}\sim\mathrm{DP}(\alpha_0,\bbeta)}$. We place the same prior over~$\pi$.  The base measure in this case is~$\bbeta$, a countably infinite vector of stick-breaking weights, that serves as the mean of the DP prior over the rows of~$\matr P$. The concentration parameter,~$\alpha_0$, governs how concentrated the rows are about the mean. Since the base measure~$\bbeta$ is discrete, each row of~$\matr P$ will be able to ``see'' the same set of states. By contrast, if we remove the HDP prior and treat each row of~$P$ as an independent draw from a DP with base measure~$H$, each row would see a disjoint set of states with probability one. In other words, the hierarchical prior is required to provide a discrete (but countably infinite) set of latent states for the HMM.

\subsection{Overdispersed Poisson Model}

An interesting consequence of this Bayesian model is that it naturally leads to a distribution of spike counts that is overdispersed relative to simple Poisson model, a feature that has been observed in neural recordings \citep{Goris14}. Recent work has explored the negative binomial distribution as an alternative to Poisson model, since its two parameters allow for Fano factors greater than one.  The negative binomial (NB) distribution can also be seen as a continuous mixture of Poisson distributions (i.e., a compound probability distribution) where the mixing distribution of the Poisson rate is a gamma distribution \citep{Gelman13}. In other words, the NB distribution is viewed as a gamma-Poisson (mixture distribution): 
a $\texttt{Poisson}(\lambda)$ distribution whose rate  $\lambda$ is itself a gamma random variable. In our case, the gamma prior over firing rates leads to a negative binomial marginal distribution over~$y_{c,t}$.

Though the marginal spike count at a particular time~$t$ may be marginally distributed according to a negative binomial distribution, it is not necessarily true that a sequence of time bins,~$\by_{1:T}$, will be i.i.d. negative binomials. This arises from the correlations induced by state transition matrix. Instead,~$\by_{1:T}$ will follow a finite mixture of Poisson distributions, with one component for each latent state. The mixture will be weighted by the marginal probability of the corresponding latent state. However, as the number of visited states grows, and the marginal probability of latent states becomes more uniform, the resulting marginal distribution over the sequence of spike counts inherits the over dispersed nature of the negative binomial distribution. This is particularly true of a HDP-HMM with a high concentration.

 \subsection{Markov Chain Monte Carlo (MCMC) Inference}

Several MCMC-based inference methods have been developed for the HDP-HMM \citep{Beal02,Teh06,van08}. Some of these previous works use a collapsed Gibbs sampler in which the transition matrix~$P$ and the observation parameters~$\bLambda$ are integrated out \citep{van08, Teh06}. In this work, however, we use a ``weak limit'' approximation in which the DP prior is approximated with a symmetric Dirichlet prior. Specifically, we let
\begin{align}
\bbeta \given \gamma &\sim \texttt{Dir}(\gamma/M, \ldots, \gamma/M), \\
\nonumber \bpi \given \alpha_0, \bbeta &\sim \texttt{Dir}(\alpha_0 \beta_1, \ldots, \alpha_0 \beta_M), \\
\nonumber \bP_{i,:} \given \alpha_0, \bbeta &\sim \texttt{Dir}(\alpha_0 \beta_1, \ldots, \alpha_0 \beta_M).
\end{align}
where $M$ denotes a truncation level for approximating the infinity (which is different from  $m$ in the finite-state setting).
 It can be shown that this prior will weakly converge to the DP prior as the dimensionality of the Dirichlet distribution approaches infinity \citep{Johnson14, Ishwaran02}. With this approximation we can capitalize on forward-backward sampling algorithms to jointly update the latent states~$\mathcal{S}$.  

Previous work has typically been presented with Gaussian or multinomial likelihood models, with the acknowledgement that the same methods work with any exponential family likelihood  when the base measure,~$H$ is a conjugate prior.
Here we present the Gibbs sampling algorithm of~\citet{Teh06} for the HDP-HMM with independent Poisson observations for each cell, and derive Hamiltonian Monte Carlo (HMC) transitions to sample the cell-specific hyperparameters of the firing rate priors. 

We begin by defining Gibbs updates for the neuronal firing rates~$\bLambda$. Since we are using gamma priors with independent Poisson observations, the model is fully conjugate and simple Gibbs updates suffice. We have,
\begin{align}
\lambda_{c,i} \given \by, \mathcal{S} &\sim \texttt{Gamma}\left(\alpha_c^0 + \sum_{t=1}^T y_{c,t} \bbI[S_t=i], \;\beta_c^0 + \sum_{t=1}^T \bbI[S_t=i]\right).
\end{align}

Under the weak limit approximation the priors on~$\bP_{i,:}$ and~$\bpi$ reduce to Dirichlet distributions, which are also conjugate with the finite HMM. Hence we can derive conjugate Gibbs updates for these parameters as well. They take the form:
\begin{align}
\bpi \given \alpha_0, \bbeta &\sim \texttt{Dir}\left(\alpha_0\bbeta + 1_{S_1}\right),\\
\nonumber \bP_{i,:} \given \alpha_0, \bbeta &\sim \texttt{Dir}\left(\alpha_0\bbeta + \bn_{i}\right),\\
\nonumber n_{i,j}&=\sum_{t=1}^{T-1} \bbI[S_t=i,S_{t+1}=j],
\end{align}
where~$1_j$ is a unit vector with a one in the~$j$-th entry.

Conditioned upon the firing rates, the initial state distribution, and the transition matrix, we can jointly update the latent states of the HDP-HMM using a {\em  forward message passing, backward sampling} algorithm. Details can be found in \citet{Johnson14b}; the intuition is that in the backward pass of the algorithm, we have a conditional distribution over~$S_t$ given~$S_{t+1:T}$. We can iteratively sample from these distributions as we go backward in time to generate a full sample from~${p(\mathcal{S} \given \bP, \bpi, \bLambda)}$. Jointly sampling these latent states allows us to avoid issues with mixing when individually sampling states that are highly correlated with one another.

Finally, the Dirichlet parameters~$\bbeta$ and the concentration parameters~$\alpha_0$ and~$\gamma$ can be updated using standard techniques. For more information, see~\citet{Teh06}. A single iteration of the final algorithm consists of an update for each parameter of the model. The aforementioned updates are based upon previous work; one novel direction that we explore in this work is the sampling of the hyperparameters of the gamma firing rate priors.

\subsubsection{Setting Firing Rate Hyperparameters} 
We consider two approaches to setting the hyperparameters of the gamma priors for Poisson firing rates, namely,~${\{\alpha_c^0, \beta_c^0\}}$ for the $c$-th cell.  Unfortunately they do not have a simple conjugate prior, but we can resort to other Bayesian techniques. First, these parameters may be estimated using an empirical Bayesian (EB) procedure, that is, by maximizing the marginal likelihood of the spike counts. For each cell, this may be easily done using standard maximum likelihood estimation for the negative binomial model.  Alternatively, we can place a weak prior and sample the hyperparameters using HMC. For simplicity, we use an improper, uniform prior on~$\log \alpha_c^0$ and~$\log \beta_c^0$. 

To implement HMC we must have access to both the log probability of the parameters as well as its gradient. Since both parameters are restricted to be positive, we instead reparameterize the problem in terms of their logs. For cell~$c$, the conditional log probability equal to,
\begin{align}
\mathcal{L} &= \log p(\log \alpha_c^0, \log \beta_c^0 \given \bLambda_{c,:}) \\
\nonumber &= \sum_{i=1}^m \log p(\lambda_{c,i} \given \alpha_c^0, \beta_c^0) + \text{const.} \\
\nonumber &= \sum_{i=1}^m \alpha_c^0 \log \beta_c^0 - \log \Gamma(\alpha_c^0) + (\alpha_c^0 - 1) \log \lambda_{c,i} - \beta_c^0 \lambda_{c,i}.
\end{align}
Taking gradients with respect to both parameters yields,
\begin{align}
\frac{\partial \mathcal{L}}{\partial \log \alpha_c^0} &= \sum_{i=1}^m \left[ \log \beta_c^0 -\Psi(\alpha_c^0) + \log \lambda_{c,i} \right] \times \alpha_c^0, \\
\nonumber \frac{\partial \mathcal{L}}{\partial \log \beta_c^0} &= \sum_{i=1}^m \left[ \frac{\alpha_c^0}{\beta_c^0}  - \lambda_{c,i} \right] \times \beta_c^0.
\end{align}

With the parameter and hyperparameter inference complete, we evaluate the performance of our algorithm in terms of its predictive log likelihood on held out test data. We approximate the predictive log likelihood with samples from the posterior distribution generated by our MCMC algorithm. That is,
\begin{align}
\log p(\by_{test} \given \by_{1:T}) &= \log \sum_{\mathcal{S}_{test}} \int_{\bTheta} p\left(\by_{test}, \mathcal{S}_{test} \given \btheta \right) \; p\left(\btheta \given \by_{train} \right) \mathrm{d}\btheta, \\
\nonumber &\approx \log \frac{1}{N} \sum_{n=1}^N \sum_{\mathcal{S}_{test}} p\left(\by_{test},  \mathcal{S}_{test} \given \btheta_n\right),
\end{align}
where~${\btheta=(\bLambda, \bP, \bpi)}$ and~${\btheta_n \sim p(\btheta \given \by_{train})}$. The summation over latent state sequences for the test data is performed with the message passing algorithm for HMMs.

\subsection{Variational Bayes (VB) Inference}

We build upon our previous work \citep{Chen12a,Chen14} as well as the recent work of \citet{Johnson14} to develop a variational inference algorithm for fitting the HDP-HMM to hippocampal spike trains. Our objective is to approximate the posterior distribution of the HDP-HMM with a distribution from a more tractable family. As usual, we choose a factorized approximation that allows for tractable optimization of the parameters of the variational model. Specifically, we let,
\begin{align}
p(\mathcal{S}, \bLambda, \bP, \bpi, \bbeta \given \by_{1:T}) &\approx q(\mathcal{S}) q(\bLambda) q(\bP) q(\bpi) q(\bbeta).
\end{align}

Since the independent Poisson observations are conjugate with the gamma firing rate prior distributions, choosing a set of independent gamma distributions for~$q(\bLambda)$ allows for simple variational updates.
\begin{align}
q(\bLambda) &= \prod_{i=1}^M \prod_{c=1}^C \texttt{Gamma}\left(\tilde{\alpha}_{c,i},\, \tilde{\beta}_{c,i} \right), \\
\nonumber \tilde{\alpha}_{c,i} &\leftarrow \alpha_c^{0} + \sum_{t=1}^T y_{c,t} \mathbb{E}_q[\bbI[S_t=i]], \\
\nonumber \tilde{\beta}_{c,i} &\leftarrow\beta_c^{0} + \sum_{t=1}^T \mathbb{E}_q[\bbI[S_t=i]].
\end{align}

Following \citet{Johnson14}, we use a ``direct assignment'' truncation for the HDP \citep{Bryant12, Liang07}. In this scheme, a truncation level~$M$ is chosen a priori and~$q(\mathcal{S})$ is limited to support only states~${S_t\in\{1,\ldots,M\}}$. The advantage of this approximation is that conjugacy is retained with~$\bLambda$,~$\bP$, and~$\bpi$, and the variational approximation~$q(\mathcal{S})$ reduces to:
\begin{align}
q(\mathcal{S}) &= \texttt{HMM}(\tilde{\bP}, \tilde{\bpi}, \tilde{\bLambda}), \\
\nonumber \tilde{\bP} &= \exp\left\{\mathbb{E}_q[\ln \bP]\right\}, \\
\nonumber \tilde{\bpi} &= \exp\left\{\mathbb{E}_q[\ln \bpi]\right\}, \\
\nonumber \tilde{\bLambda} &= \exp\left\{\mathbb{E}_q[\ln p(\by \given \bLambda)]\right\}.
\end{align}
Expectations~${\mathbb{E}_q[S_t=i]}$ can then be computed using standard message passing algorithms for HMMs.

With the direct assignment truncation, the variational factors for~$\bP_{i,:}$ and~$\bpi$ take on Dirichlet priors. Unlike in the finite HMM, however, these Dirichlet priors are now over~${M+1}$ dimensions since the final dimension accounts for all states~${i>M}$. Under the HDP prior we had~${P_{i,:}\sim \texttt{DP}(\alpha_0 \cdot \bbeta)}$, and under the truncation the DP parameter becomes~$\alpha_0 \cdot \bbeta_{1:M+1}$. Again, leveraging the conjugacy of the model, we arrive at the following variational updates:
\begin{align}
q(\bP) &= \prod_{i=1}^M \mathrm{Dir}(\tilde{\balpha}_P^{(i)}), \\
\nonumber (\tilde{\balpha}_P^{(i)})_j &\leftarrow \alpha_0 \beta_j + \mathbb{E}_q[\bbI[S_t=i]\cdot\bbI[S_{t+1}=j]].
\end{align}
We use an analogous update for~$\bpi$.

The principal drawback of the direct assignment truncation is that the prior for~$\bbeta$ is no longer conjugate. This could be avoided with the fully conjugate approach of \cite{Hoffman13}, however, this results in extra bookkeeping and the duplication of states. Instead, following \cite{Johnson14, Bryant12, Liang07}, we use a point estimate for this parameter by setting~${q(\bbeta)=\delta_{\bbeta^*}}$ and use gradient ascent with backtracking line search to update this parameter during inference.

There are a number of hyperparameters to set for the variational approach as well. The hyperparameters~$\alpha_c^0$ and~$\beta_c^0$ of gamma prior on firing rates can be set with empirical Bayes, as above. We resort to cross validation in order to set the Dirichlet parameter~$\alpha_0$ and the GEM parameter~$\gamma$. 

Finally, in order to compute predictive log likelihoods on held out test data, we draw multiple samples~$\{(\bLambda, \mathcal{S}, \bP, \bpi, \bbeta)_n\}_{n=1}^N$ from the variational posterior and approximate the predictive log likelihood as,
\begin{align}
\ln p(\by_{test} \given \by_{1:T}) &\approx \ln \mathbb{E}_q\left[p(\by_{test} \given \mathcal{S}, \bLambda,  \bP, \bpi, \bbeta)\right] \\
\nonumber &\approx \ln \frac{1}{N} \sum_{n=1}^N p(\by_{test} \given (\mathcal{S}, \bLambda, \bP, \bpi, \bbeta)_n).
\end{align}

\section{Results}
The inference algorithms were implemented based upon the PyHSMM framework of~\citet{Johnson14b}. The codebase was written in Python with \texttt{C} offloads for the message passing algorithms.  We extended the codebase to perform hyperparameter inference using the methods described above, and expanded it to tailor to neural spike train analysis. Our code is publicly available ({\url{https://github.com/slinderman}}).

\begin{figure}
\centering
\begin{subfigure}[t]{4.in}
\includegraphics[width=\textwidth]{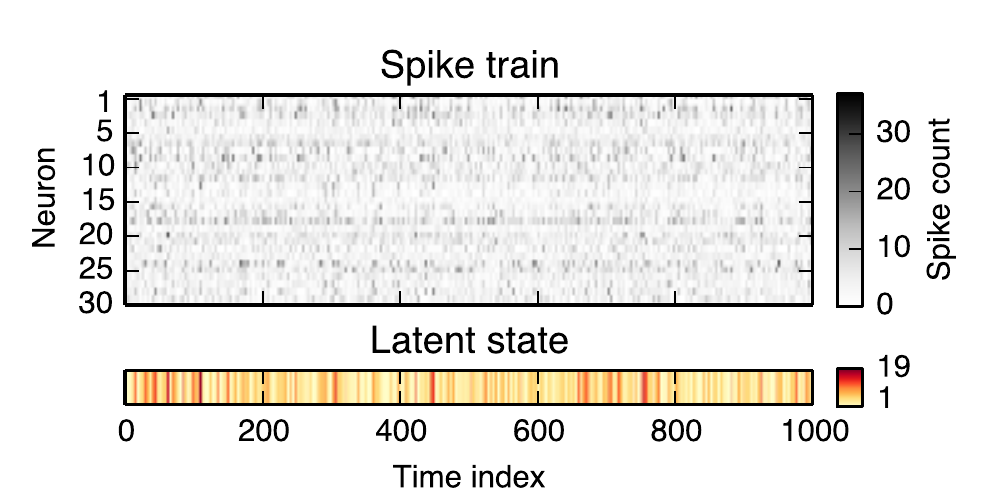}
\end{subfigure}
\\
\vspace{-.2in}
\begin{subfigure}[t]{2in}
\includegraphics[width=\textwidth]{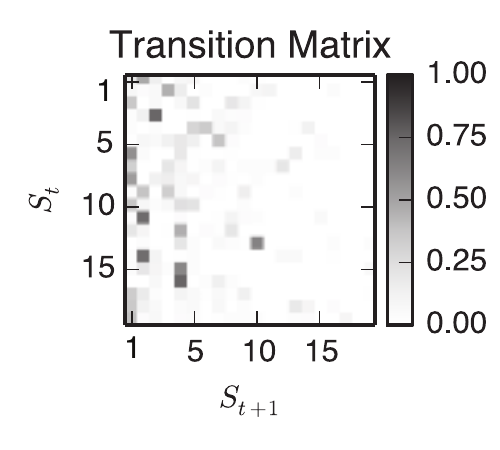}
\end{subfigure}
~
\begin{subfigure}[t]{2.5in}
\includegraphics[width=\textwidth]{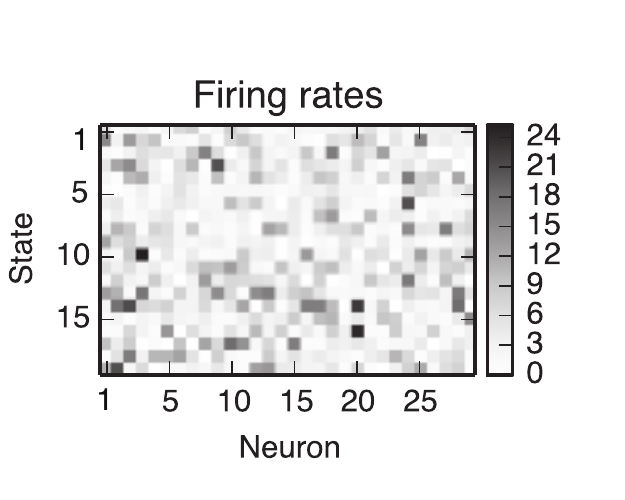}
\end{subfigure}
\\
\vspace{-.2in}
\caption{An example of a synthetic dataset drawn from an HDP-HMM.  From the observed spike trains (top), we infer the latent state sequence (middle), the transition matrix~$\matr P$ (lower left), and the firing rate vectors~$\blambda_i$ (rows in lower right) specific to each state.}
\label{fig1}
\end{figure}

 \subsection{Synthetic Data}

 First, we simulate synthetic spike count data using an HDP-HMM with ${C=30}$ neurons,~${T=1000}$ time bins, and Dirichlet concentration parameters~$\alpha_0=4.0$ and~$\gamma=8.0$. These yield state sequences that tend to visit~15-30 states. All of neuronal firing rate parameters are drawn from a gamma distribution: $\texttt{Gamma}(1, 0.2)$ (with mean 5 and standard deviation 5). 
 
 An example of one such synthetic dataset is shown in Fig.~\ref{fig1}. The states have been ordered according to their occupancy (i.e., how many times they are visited during the simulation), such that the columns of the transition matrix exhibit a decrease in probability as the incoming state number,~${S}_{t+1}$, increases. This is a characteristic of the HDP-HMM, and indicates the tendency of the model to reuse states with high occupancy.

We compare five combinations of model,  inference algorithm, and hyperparameter selection approaches: (i) HMM with the correct number of states, fit by MCMC and HMC; (ii) HMM with the correct number of states, fit by VB with hyperparameters set by EB; (iii) HDP-HMM fit by MCMC with HMC for hyperparameter selection; (iv) HDP-HMM fit by MCMC with EB for hyperparameters; and (v) HDP-HMM fit by VB with hyperparameters set by EB. For the MCMC methods, we set gamma priors over~$\alpha_0$ and~$\gamma$ and use Gibbs sampler as described above; for the VB methods, we set~$\alpha_0$ and~$\gamma$ to their true values. Alternatively, they can be selected by cross validation. We set both the weak limit approximation for MCMC and the direct assignment truncation level for VB to~${M=80}$.

We collect 300 samples from the MCMC algorithms and use the last 50 for computing predictive log likelihoods. For visualization, we use the final sample to extract the transition matrix and the firing rates. The number of samples and the amount of burn-in iterations were chosen by examining the log probability and parameter traces for convergence. It is found that the MCMC algorithm converges within tens of iterations. For further convergence diagnosis of a single Gibbs chain, one may use the autocorrelation tools suggested in \citep{RafteryLewis92,Cowles96}.

We run the VB algorithm for 100 steps to guarantee convergence of the variational lower bound. Again, this is assessed by examining the variational lower bound and is found to converge to a local maxima within tens of iterations.

\begin{figure}
\center  
\begin{subfigure}[t]{2.5in}
\begin{subfigure}[t]{1.2in}
\includegraphics[width=\textwidth]{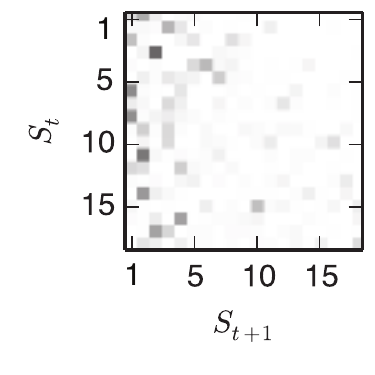}
\end{subfigure}
~
\begin{subfigure}[t]{1.2in}
\includegraphics[width=\textwidth]{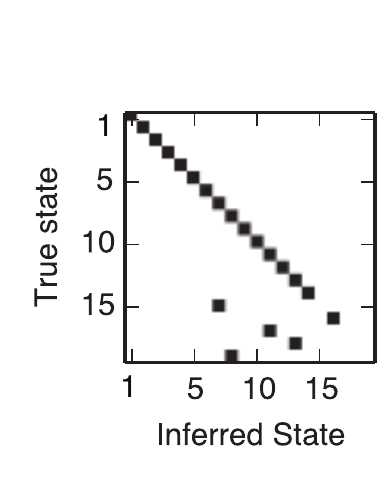}
\end{subfigure}
\caption{HMM (MCMC+HMC)}
\end{subfigure}
~
\begin{subfigure}[t]{2.5in}
\begin{subfigure}[t]{1.2in}
\includegraphics[width=\textwidth]{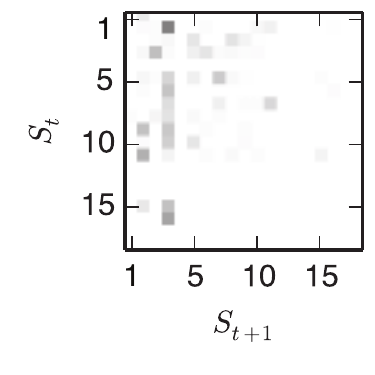}
\end{subfigure}
~
\begin{subfigure}[t]{1.2in}
\includegraphics[width=\textwidth]{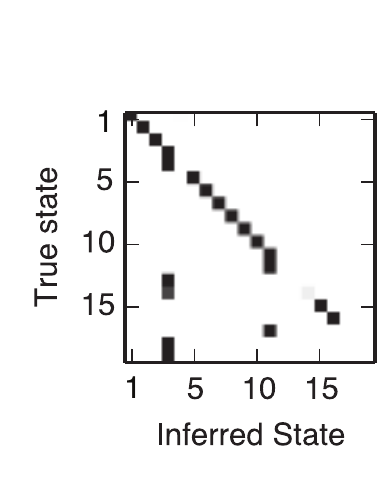}
\end{subfigure}
\caption{HMM (VB)}
\end{subfigure}
\\
\begin{subfigure}[t]{2.5in}
\begin{subfigure}[t]{1.2in}
\includegraphics[width=\textwidth]{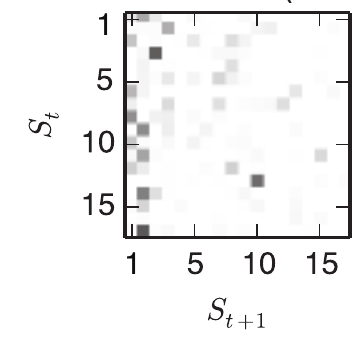}
\end{subfigure}
~
\hspace{-.1in}
\begin{subfigure}[t]{1.2in}
\includegraphics[width=\textwidth]{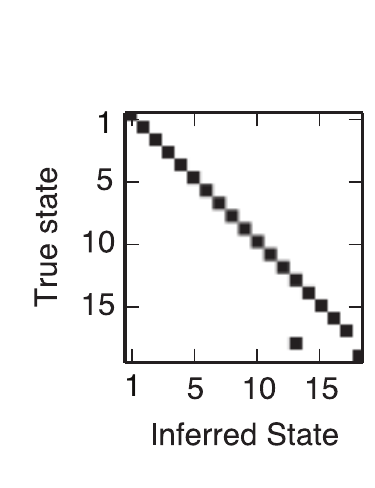}
\end{subfigure}
\caption{HDP-HMM (MCMC+HMC)}
\end{subfigure}
~
\begin{subfigure}[t]{2.5in}
\begin{subfigure}[t]{1.2in}
\includegraphics[width=\textwidth]{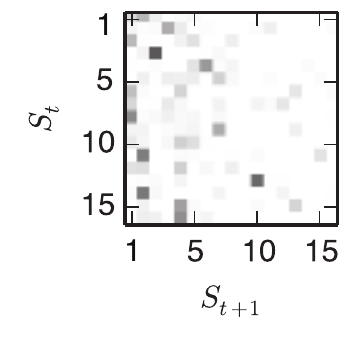}
\end{subfigure}
~
\hspace{-.1in}
\begin{subfigure}[t]{1.2in}
\includegraphics[width=\textwidth]{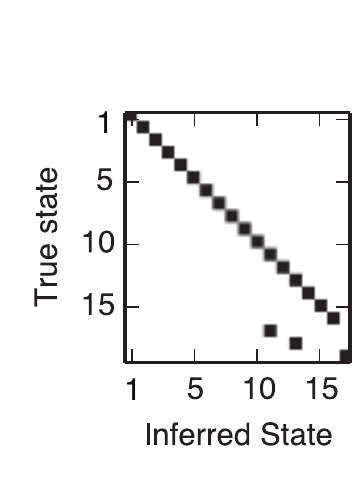}
\end{subfigure}
\caption{HDP-HMM (MCMC+EB)}
\end{subfigure}
\\
\begin{subfigure}[t]{3.1in}
\begin{subfigure}[t]{1.45in}
\includegraphics[width=\textwidth]{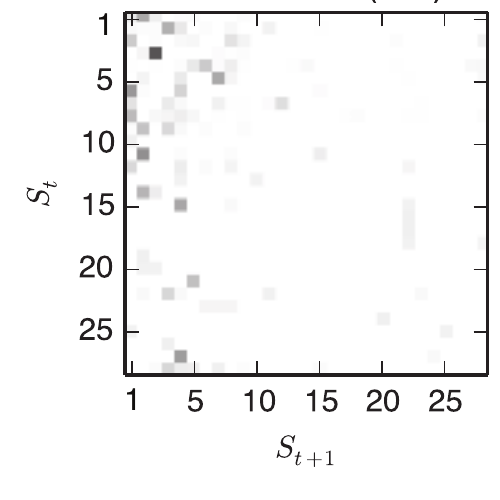}
\end{subfigure}
~
\hspace{-.1in}
\begin{subfigure}[t]{1.55in}
\includegraphics[width=\textwidth]{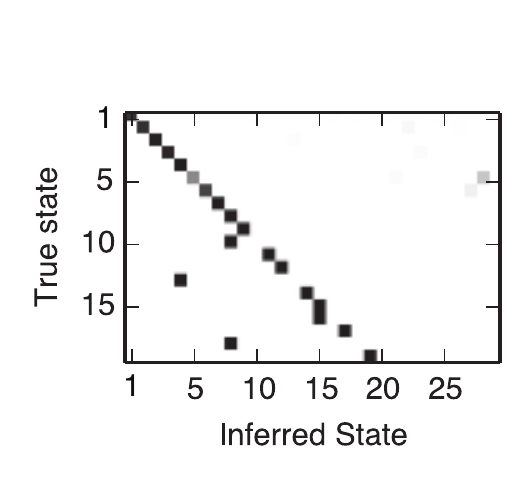}
\end{subfigure}
\caption{HDP-HMM (VB)}
\end{subfigure}
\caption{A comparison of inferred state transition matrices and their alignment with the true transition matrix using five combinations of model and inference algorithm. The true state transition matrix is shown in Fig.~\ref{fig1}, and is drawn from an HDP-HMM. We find that the MCMC-based inference algorithms yield state estimates that have better correspondence with the true states. The HDP-HMM fit using MCMC combined with HMC for hyperparameter sampling achieves the best match, as evidenced by the nearly one-to-one mapping between true and inferred states (a diagonal matrix indicates a perfect match). The VB algorithms find state sequences with many-to-many maps between true and inferred states. A complete description of the models and inference algorithms is given in the main text.}
\label{fig2}
\end{figure}

We use two criteria for result assessment with simulation data.  The first criterion is based on the state-state map. Ideally, if two state sequences are consistent, the scatter plot of two state sequences will have a one-to-one mapping.  The second criterion is the model's predictive log likelihood (unit: bits/spike) on a held out sequence of~${T_{test}=200}$ time steps. We compare the predictive log likelihood to that of a set of independent Poisson processes. Their rates and the corresponding predictive log likelihood are given by,
\begin{align}
\hat{\lambda}_c&=\frac{1}{T_{train}}\sum_{t=1}^{T_{train}} y_{c,t}, \\
\log p(\by_{test} \given \by_{train}) &= \sum_{c=1}^C \left[ -T_{test} \hat{\lambda}_c + \sum_{t=1}^{T_{test}} y_{c,t} \log \hat{\lambda}_c  \right].
\end{align}
The improvement obtained by a model is measured in bits, and is normalized by the number of spikes in the test dataset in order to obtain comparable units for each of the test datasets.

Figure~\ref{fig2} shows the inferred state transition matrices (left half of each panel) and the corresponding state-state map between the true and inferred states (right half) for each of the model, inference algorithm, and initialization combinations. If the mapping is one-to-one, the map will be square and each row will have only one nonzero entry. To facilitate visualization in the presence of permutation ambiguity, we have reordered the inferred states to make the resulting map as diagonal as possible. We use a simple algorithm to match inferred states to true states based on their overlap. We begin by marking all true and inferred states as ``unmatched,'' then we find the pair of true and inferred states with the highest overlap, mark them as ``matched,'' remove them from the list of unmatched states, and repeat until either the true or inferred states have all been matched.  We use this greedy mapping to permute the inferred states for presentation. Since the HDP-HMM may infer a different number of states than are present in the true model, the map may not be a square matrix (e.g., Fig.~\ref{fig2}e). Again, the true states are ordered according to their occupancy so that true state with the lowest index is the most visited state.

Table~\ref{table1} summarizes the predictive log likelihood comparison among 10 independent runs. For 7 of the 10 datasets, the HDP-HMM fit based on MCMC and HMC performs best, though in some cases the difference between the HDP-HMM when using HMC versus EB for hyperparameter selection is miniscule. In the cases where the HDP-HMM (MCMC+HMC) is not the best, its performance differs by at most 0.001 bits/spike.


Though computation cost is often a major factor with Bayesian inference, with the optimized PyHSMM package, the models can be fit to the synthetic data in under five minutes on an Apple MacBook Air. The runtime necessarily grows the number of neurons and the truncation limit on the number of latent states. As the model complexity grows, we must also run our MCMC algorithm for more iterations, which often motivates the use of variational inference algorithms instead. Given our optimized implementation and the performance improvements yielded by MCMC, we opted for a fully-Bayesian approach using MCMC with HMC for hyperparameter sampling in our subsequent experiments.

 \begin{table}
\centering
\caption{Predictive log likelihood (bits/spike) comparison for 10 simulated datasets, measured in bits per spike improvement over a baseline of independent, homogeneous Poisson processes (the best result in each data set is marked in bold font). }
\begin{tabular}{c|ccccc}
\hline
      & \specialcell{HMM\\(MCMC+HMC)} & \specialcell{HMM\\(VB)} & \specialcell{HDP-HMM\\(MCMC+HMC)}  & \specialcell{HDP-HMM\\(MCMC+EB)}  & \specialcell{HDP-HMM\\(VB)}  \\
\hline
Dataset 1 & ${0.321}$  & ${0.286}$  & ${\mathbf{0.323}}$  & ${0.323}$  & ${0.299}$  \\
Dataset 2 & ${0.348}$  & ${0.255}$  & ${\mathbf{0.361}}$  & ${0.359}$  & ${0.350}$  \\
Dataset 3 & ${0.317}$  & ${0.294}$  & ${\mathbf{0.318}}$  & ${0.318}$  & ${0.297}$  \\
Dataset 4 & ${0.385}$  & ${0.329}$  & ${\mathbf{0.385}}$  & ${0.377}$  & ${0.353}$  \\
Dataset 5 & ${0.331}$  & ${0.310}$  & ${0.332}$  & ${0.333}$  & ${\mathbf{0.333}}$  \\
Dataset 6 & ${0.323}$  & ${0.298}$  & ${\mathbf{0.333}}$  & ${0.331}$  & ${0.329}$  \\
Dataset 7 & ${0.307}$  & ${0.274}$  & ${\mathbf{0.320}}$  & ${0.320}$  & ${0.317}$  \\
Dataset 8 & ${\mathbf{0.320}}$  & ${0.223}$  & ${0.319}$  & ${0.313}$  & ${0.319}$  \\
Dataset 9 & ${0.261}$  & ${0.251}$  & ${\mathbf{0.262}}$  & ${0.261}$  & ${0.259}$  \\
Dataset 10 & ${0.327}$  & ${0.272}$  & ${0.337}$  & ${\mathbf{0.338}}$  & ${0.335}$  \\
\hline
\end{tabular}
\label{table1}
\end{table}

 \begin{figure}
\centering
\begin{subfigure}[t]{2.40in}
\includegraphics[width=\textwidth]{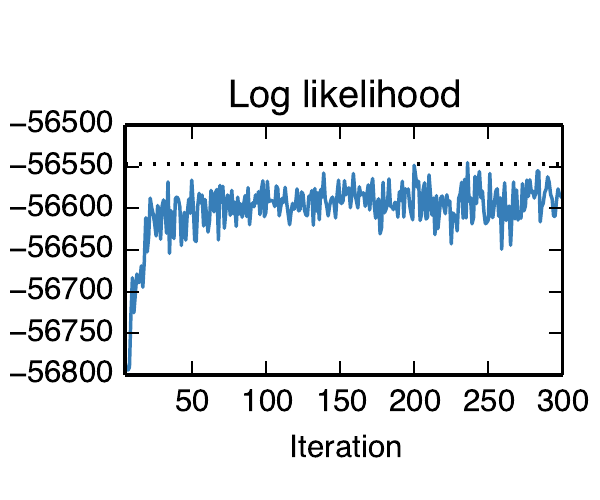}
\end{subfigure}
~
\begin{subfigure}[t]{2.40in}
\includegraphics[width=\textwidth]{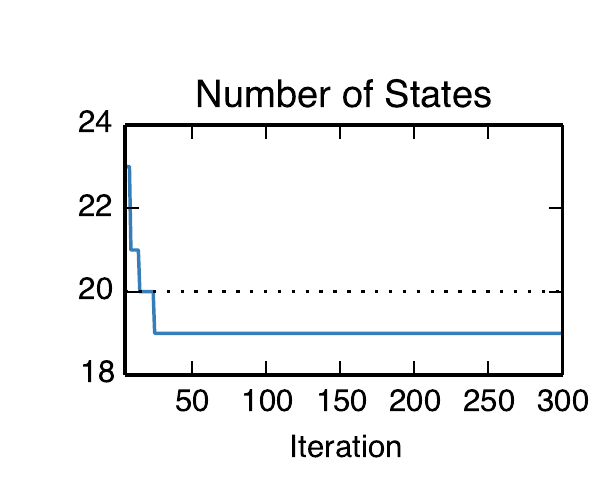}
\end{subfigure}
\\
\vspace{-.25in}
\begin{subfigure}[t]{2.40in}
\includegraphics[width=\textwidth]{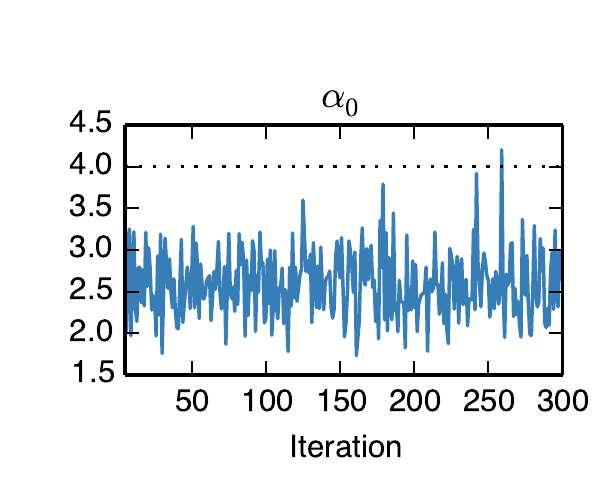}
\end{subfigure}
\begin{subfigure}[t]{2.40in}
\includegraphics[width=\textwidth]{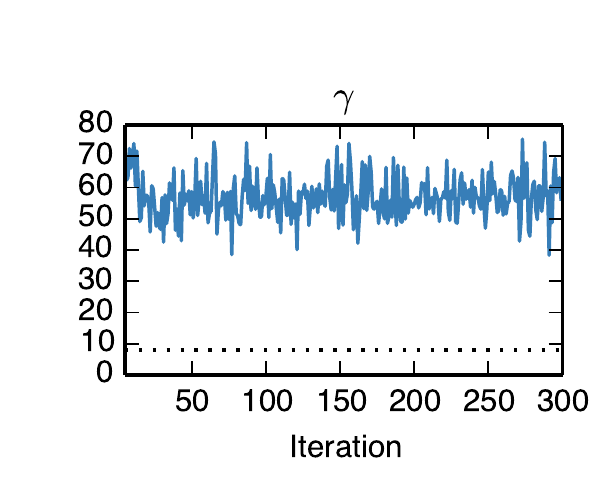}
\end{subfigure}
\caption{MCMC state trajectories for an HDP-HMM fit to the synthetic dataset shown in Fig.~\ref{fig1}. True values are shown by the dotted black lines. The first five iterations of the Markov chain are omitted since they differ greatly from the final states. The chain quickly converges to nearly the correct number of states and achieves close to the true log likelihood. }
\label{fig3}
\end{figure}

\begin{figure}
\centering
\begin{subfigure}[t]{4in}
\includegraphics[width=\textwidth]{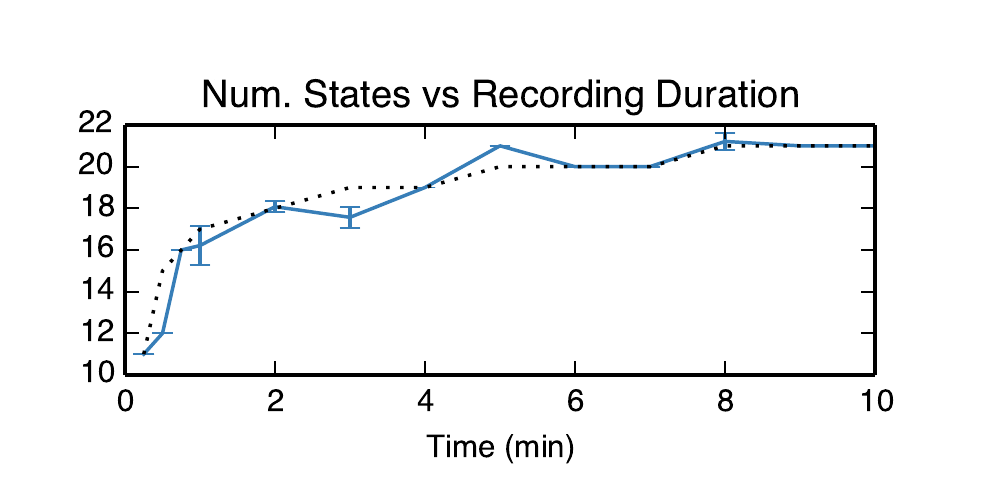}
\end{subfigure}
\vspace{-1em}
\caption{Analysis of number of inferred states as a function of length of the recording. We simulated a synthetic dataset from an HDP-HMM with 50 neurons for 10 minutes with bins of size 250 ms. At the end of the simulation, the model had visited 50 states (dotted black line). Fitting the HDP-HMM with MCMC+HMC on increasing subsets of the data shows a similar trend in number of inferred sates, plateauing at around 20-21 states given 5 minutes of data.}
\label{fig:n_states_v_duration}
\end{figure}

Figure~\ref{fig3} shows example traces from the MCMC combined with HMC algorithm for the HDP-HMM running on synthetic dataset 1. This is the same data from which Fig.~\ref{fig2} is generated. The first 5 iterations have been omitted to highlight the variation in the latter samples (the first few iterations rapidly move away from the initial conditions). We see that the log likelihood of the data rapidly converges to nearly that of the true model, and the number of states quickly converges to~$m=19$. Referring back to Fig.~\ref{fig2}, we see that this set of states has combined two of the less occupied true states into one.  Note that the nuisance parameters $\alpha_0$  and $\gamma$ do not converge to the true values within 300 iterations--- this could be due to the fact that the solution is not sensitive to these parameters or the presence of local optima.  However, even the concentration parameters are different from the true values,  they are still consistent with the inferred state transition matrix. 

Finally, we considered how the number of inferred states varies as a function of recording duration. We simulated 2400 sample points (which is equivalent to 10-min recording with 250 ms temporal bins) from an HDP-HMM with the same parameters as above. The simulation resulted in 21 states visited over the total dataset. We then fit the model with a HDP-HMM with MCMC plus HMC on increasing subset size of the data. The model was initialized with a~$\texttt{Gamma}(1.0,1.0)$ prior on~$\alpha_0$ and a~$\texttt{Gamma}(8.0,1.0)$ prior on~$\gamma$. Thus, the prior means are equal to the underlying concentration parameters. Figure~\ref{fig:n_states_v_duration} shows the number of inferred states (blue line) and the true number of states (dotted black line) as a function of recording duration. As hoped, the number of inferred states parallels the true number of states; this is indeed the case when five minutes of data are used. 
 
 \subsection{Rat Hippocampal Neuronal Ensemble Data} 

Next, we apply the proposed methods to experimental data of the rat hippocampus. 
Experiments were conducted under the supervision of the Massachusetts Institute of Technology (MIT)  Committee on Animal Care and followed the NIH guidelines.
The micro-drive arrays containing multiple tetrodes were implanted above the  right dorsal hippocampus of male Long-Evans rats. The tetrodes were slowly lowered into the brain reaching the cell layer of CA1 two to four weeks following the date of surgery. Recorded spikes were manually  clustered and sorted to obtain single units using a custom software (XClust, M.A.W.).

\begin{figure}
\centering
\begin{subfigure}[t]{\textwidth}
\centering
\includegraphics[width=1.35in]{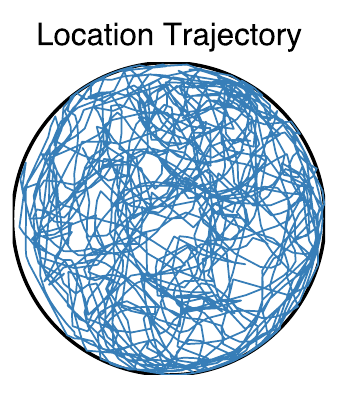}
\hspace{0.25in}
\includegraphics[width=2.25in]{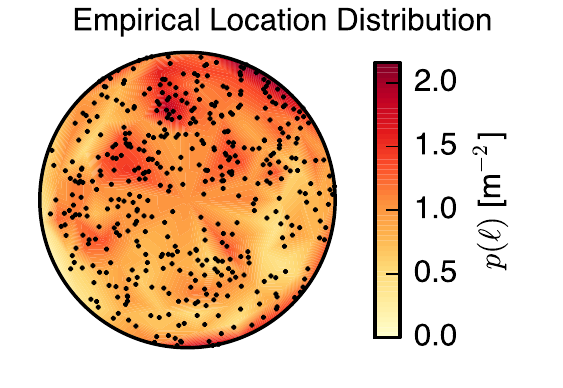}
\end{subfigure}
\caption{One rat's behavioral trajectory (left) and spatial occupancy (right) in the open field environment. }
\label{fig5}
\end{figure}

For demonstration purpose, an ensemble spike train recording of~${C=47}$ cells was collected from a single rat for a duration of 9.8 minutes.
Once   stable hippocampal units were obtained, the rat was allowed to freely forage in an approximately circular open field environment (radius: $\sim$60 cm). 
To identify the period of rodent locomotion during spatial navigation, we used a velocity threshold ($>$10 cm/s) to select the RUN epochs and merged them together. 
One animal's RUN trajectory  and spatial occupancy are shown in Fig.~\ref{fig5} (left and right panels, respectively). The empirical probability of a location,~${p(\ell)}$, is determined by dividing the arena into 220 bins of equal area (11 angular bins and 20 radial bins) and counting the fraction of time points in which the rat is in the corresponding bin. 

In the experimental data analysis, we focus on  nonparametric Bayesian inference for HDP-HMM.
For all methods, we increase the truncation level to a large value of $M=200$.  To discover the model order of the variational solutions, we use the number of states visited by the most likely state sequence under the variational posterior. The MCMC algorithms yield samples of state sequences from which the model order can be directly counted.

\begin{figure}
\centering
\begin{subfigure}[t]{1.20in}
\includegraphics[width=\textwidth]{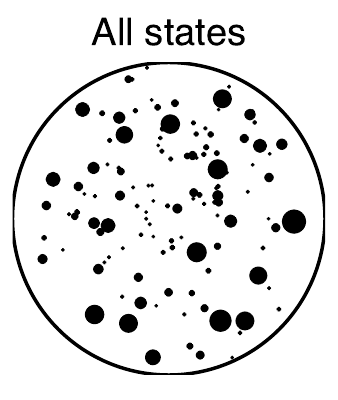}
\end{subfigure}
\hspace{-.1in}
~
\begin{subfigure}[t]{1.20in}
\includegraphics[width=\textwidth]{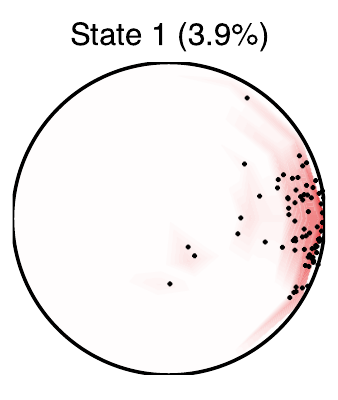}
\end{subfigure}
\hspace{-.1in}
~
\begin{subfigure}[t]{1.20in}
\includegraphics[width=\textwidth]{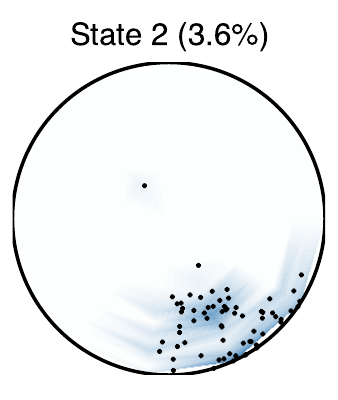}
\end{subfigure}
\hspace{-.1in}
\begin{subfigure}[t]{1.20in}
\includegraphics[width=\textwidth]{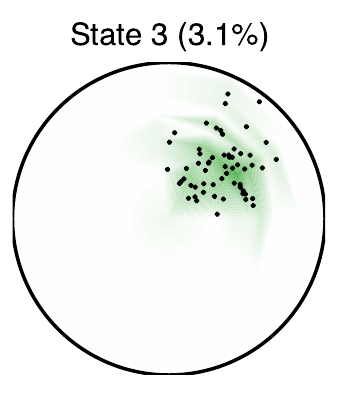}
\end{subfigure}
\\
\begin{subfigure}[t]{4.98in}
\includegraphics[width=\textwidth]{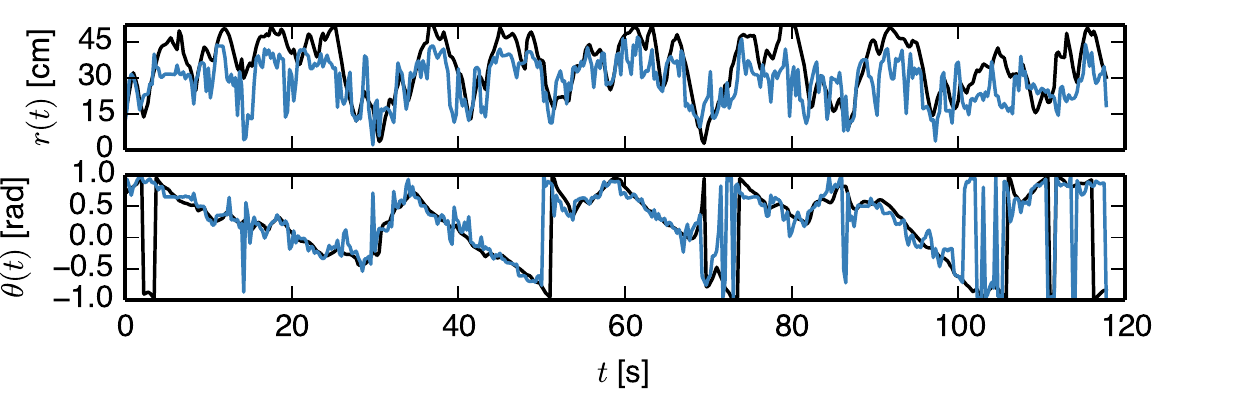}
\end{subfigure}
\caption{Estimation result from HDP-HMM (MCMC+HMC) for the rat hippocampal spike train. (Top left) Estimated state space map, where the mean value of the spatial position for each latent state is shown by a black dot. The size of the dot is proportional to the occupancy of the state.  (Top right) Probability distributions over location corresponding to the top three latent states, measured by state occupancy. The small black dots indicate the location of the animal while in that state, and are used to compute the empirical distribution over location indicated by colored shading.
(Bottom) The true trajectory (in polar coordinates) is shown in black and the reconstructed trajectory is shown in blue. For each time bin, we use the mean location of the latent states to determine an estimate of the animal's location. }
\label{fig6}
\end{figure}

For the purpose of result assessment, we plot  the state space map  (Fig.~\ref{fig6}, top left), which shows the mean value of the spatial position that each state represented. The size of the black dot is proportional to the occupancy of the state. We also plot the empirical location distribution for the top three states as measured by occupancy (Fig.~\ref{fig6}, top right). In the bottom of Fig.~\ref{fig6}, we show the estimated animal's spatial trajectories (polar coordinate) in black, along with the reconstructed location in from the HDP-HMM with MCMC and HMC in blue. To reconstruct the position, we use the mean of each latent state's location distribution weighted by the marginal probability of that state under the HDP-HMM. That is,
\begin{align}
\hat{r}_t = \sum_{i=1}^m \bar{r}_i \Pr(\mathcal{S}_t=i), \qquad
\hat{\theta}_t = \sum_{i=1}^m  \bar{\theta}_i \Pr(\mathcal{S}_t=i),
\end{align}
where~$\bar{r}_i$ and~$\bar{\theta}_i$ denote the average location of the rat while in inferred state~$i$ (corresponding to the black dots in Fig.~\ref{fig6}, top leftmost panel). Note that the animal's position is not used in model inference, only during result assessment. In the illustrated example (HDP-HMM with MCMC+HMC), the mean reconstruction error in Euclidean distance is 9.07 cm. 

\begin{figure}
\centering
\begin{subfigure}[t]{2.40in}
\includegraphics[width=\textwidth]{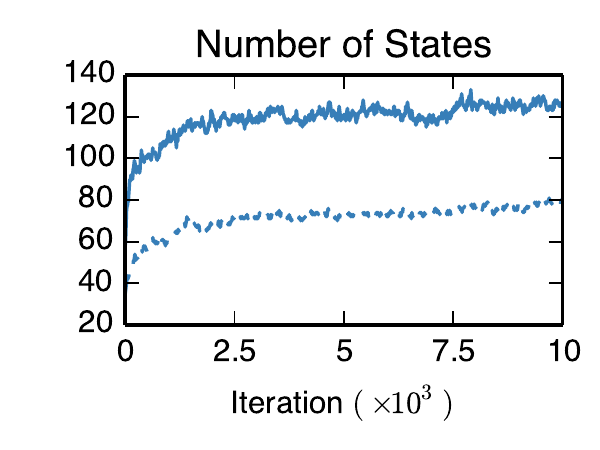}
\end{subfigure}
~
\begin{subfigure}[t]{2.40in}
\includegraphics[width=\textwidth]{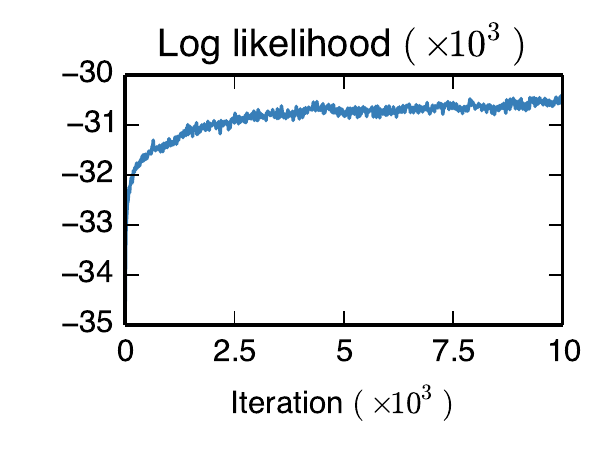}
\end{subfigure}
\\
\vspace{-.25in}
\begin{subfigure}[t]{2.40in}
\includegraphics[width=\textwidth]{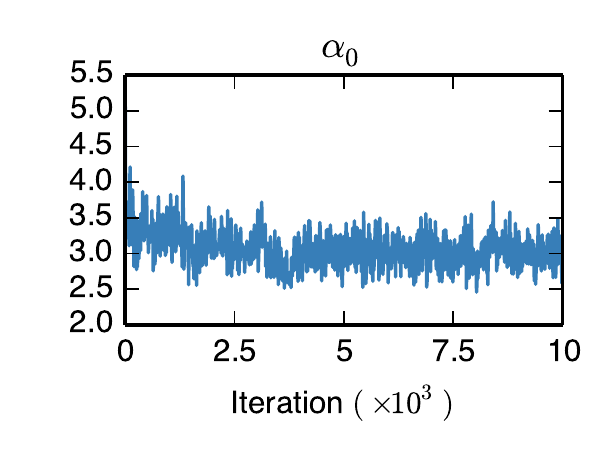}
\end{subfigure}
\begin{subfigure}[t]{2.40in}
\includegraphics[width=\textwidth]{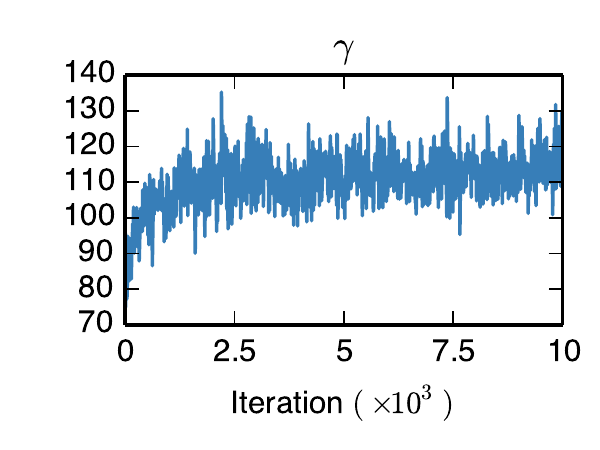}
\end{subfigure}
\\
\begin{subfigure}[b]{2.40in}
\includegraphics[width=\textwidth]{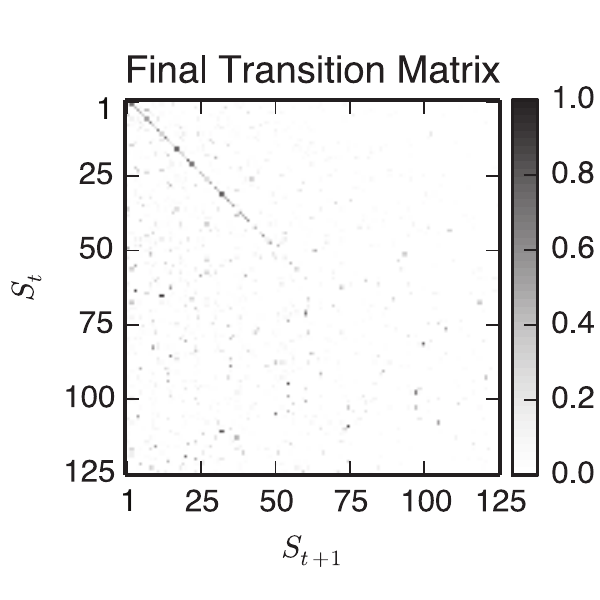}
\end{subfigure}
~
\begin{subfigure}[b]{2.40in}
\includegraphics[width=\textwidth]{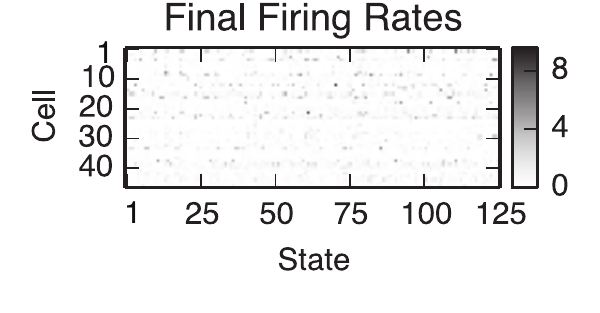}
\end{subfigure}
\caption{Estimation result from HDP-HMM (MCMC+HMC) for the rat hippocampal data. The total number of states (solid blue) slowly increases as states are allocated for a small number of time bins. The number of states accounting for 95\% of the data (dashed line) converges after 7500 iterations. The log likelihood of the training data grows consistently as highly specific states are added. The concentration parameters,~$\alpha_0$ and~$\gamma$ converge after 2500 iterations. In the bottom we show the final state transition matrix and firing rate samples drawn from the last iteration.
}
\label{fig7}
\end{figure}

\begin{figure}
\centering
\begin{subfigure}[t]{1.20in}
\includegraphics[width=\textwidth]{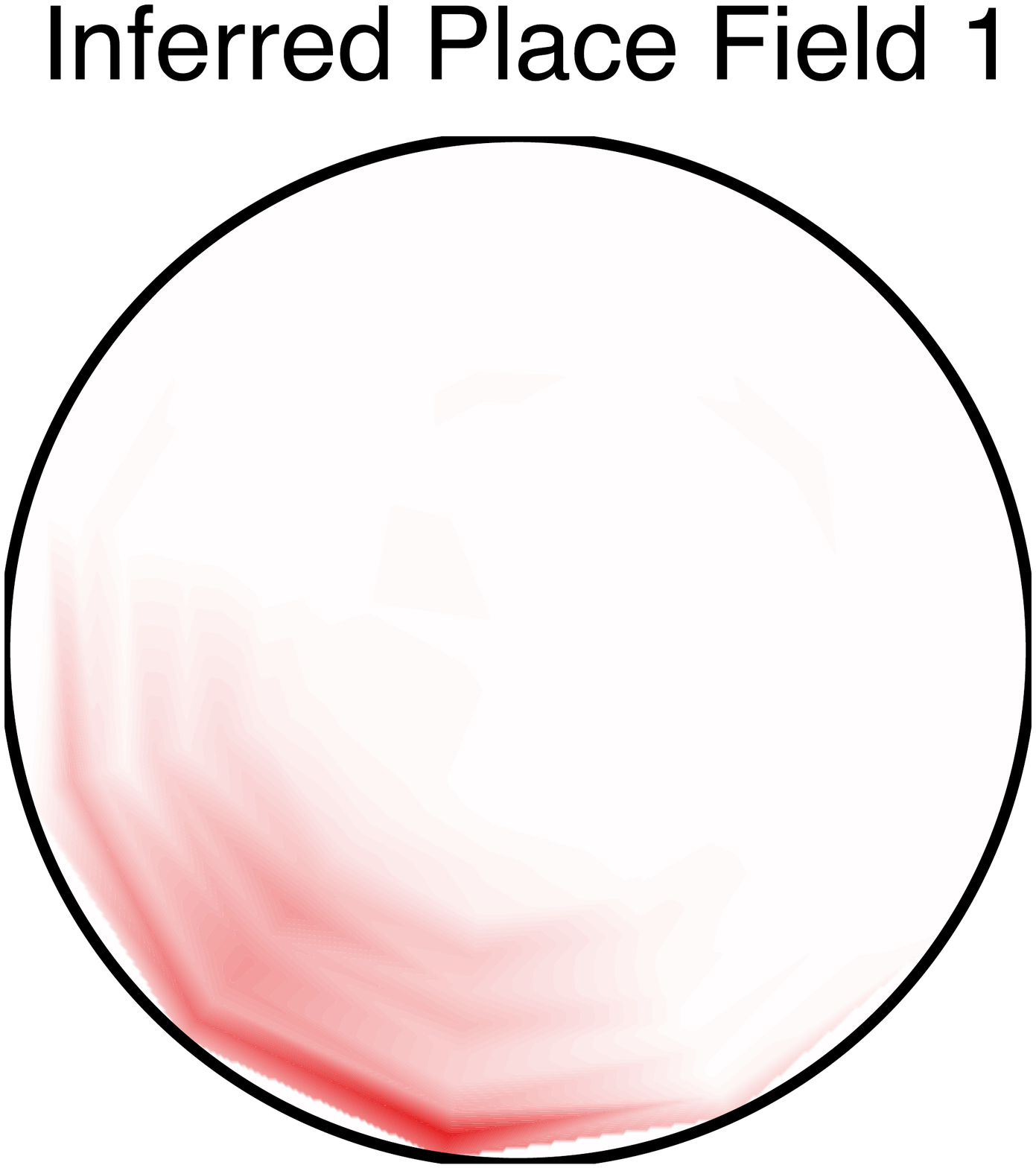}
\end{subfigure}
\hspace{-.1in}
~
\begin{subfigure}[t]{1.20in}
\includegraphics[width=\textwidth]{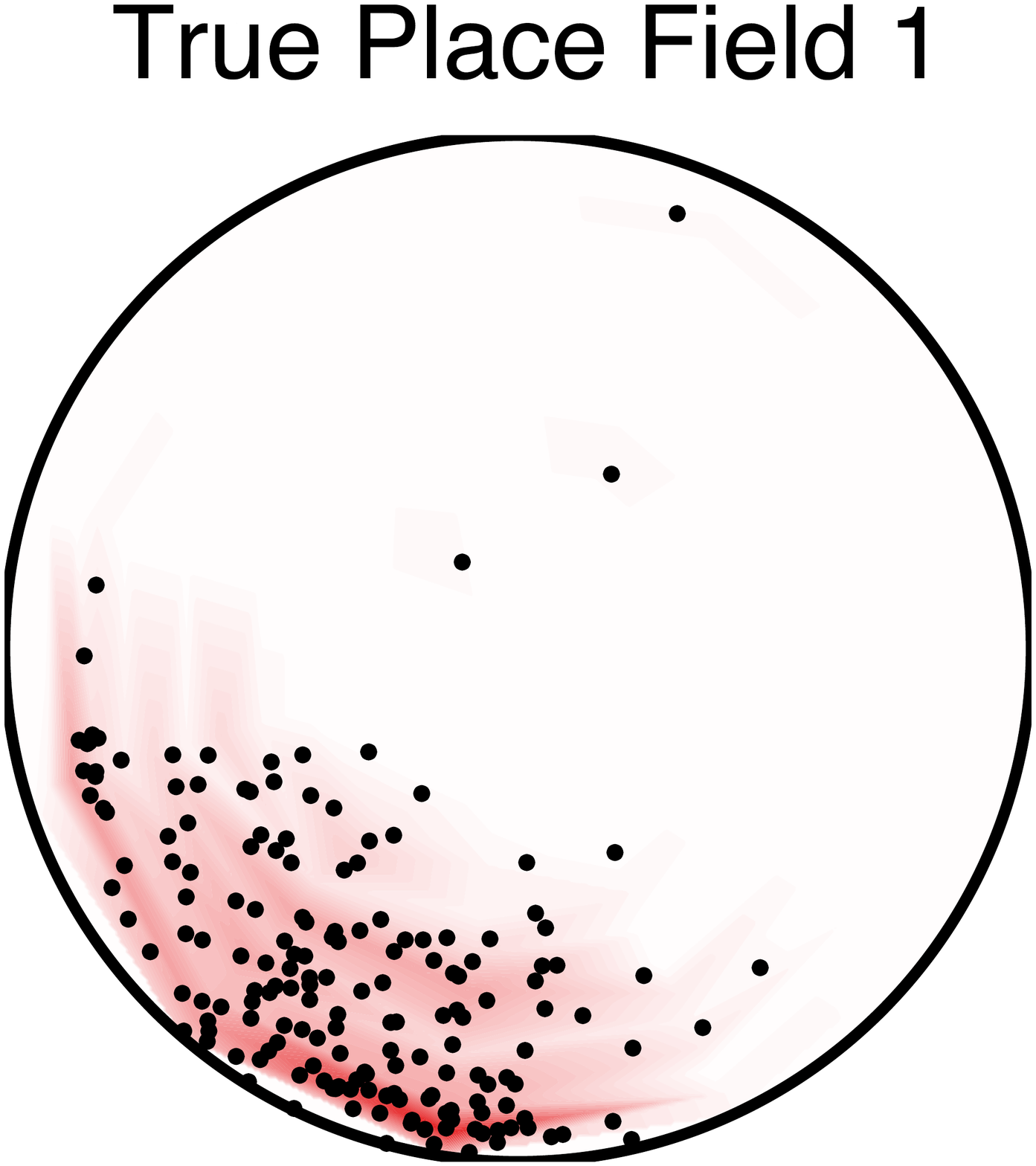}
\end{subfigure}
\hspace{-.1in}
~
\begin{subfigure}[t]{1.20in}
\includegraphics[width=\textwidth]{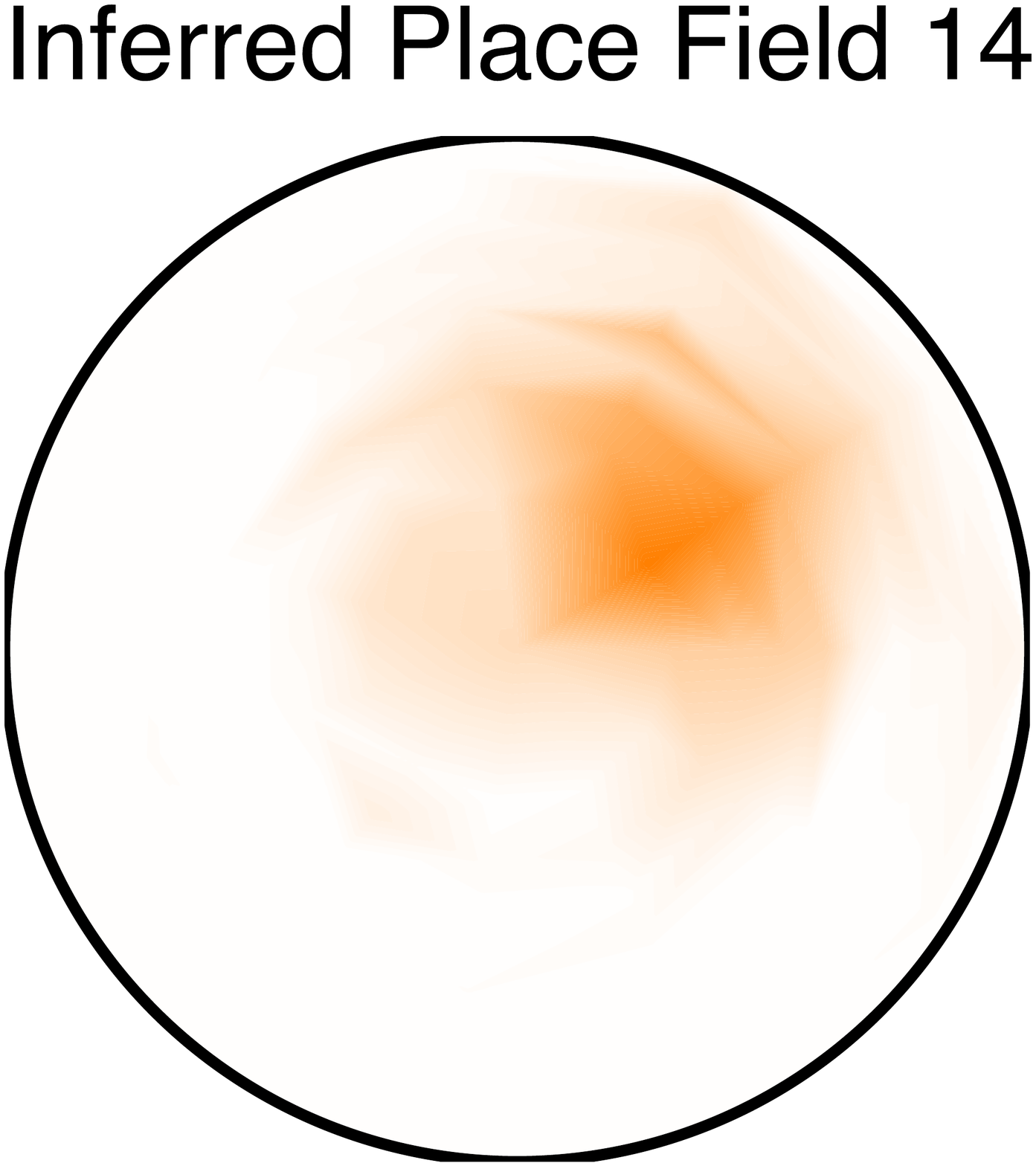}
\end{subfigure}
\hspace{-.1in}
~
\begin{subfigure}[t]{1.20in}
\includegraphics[width=\textwidth]{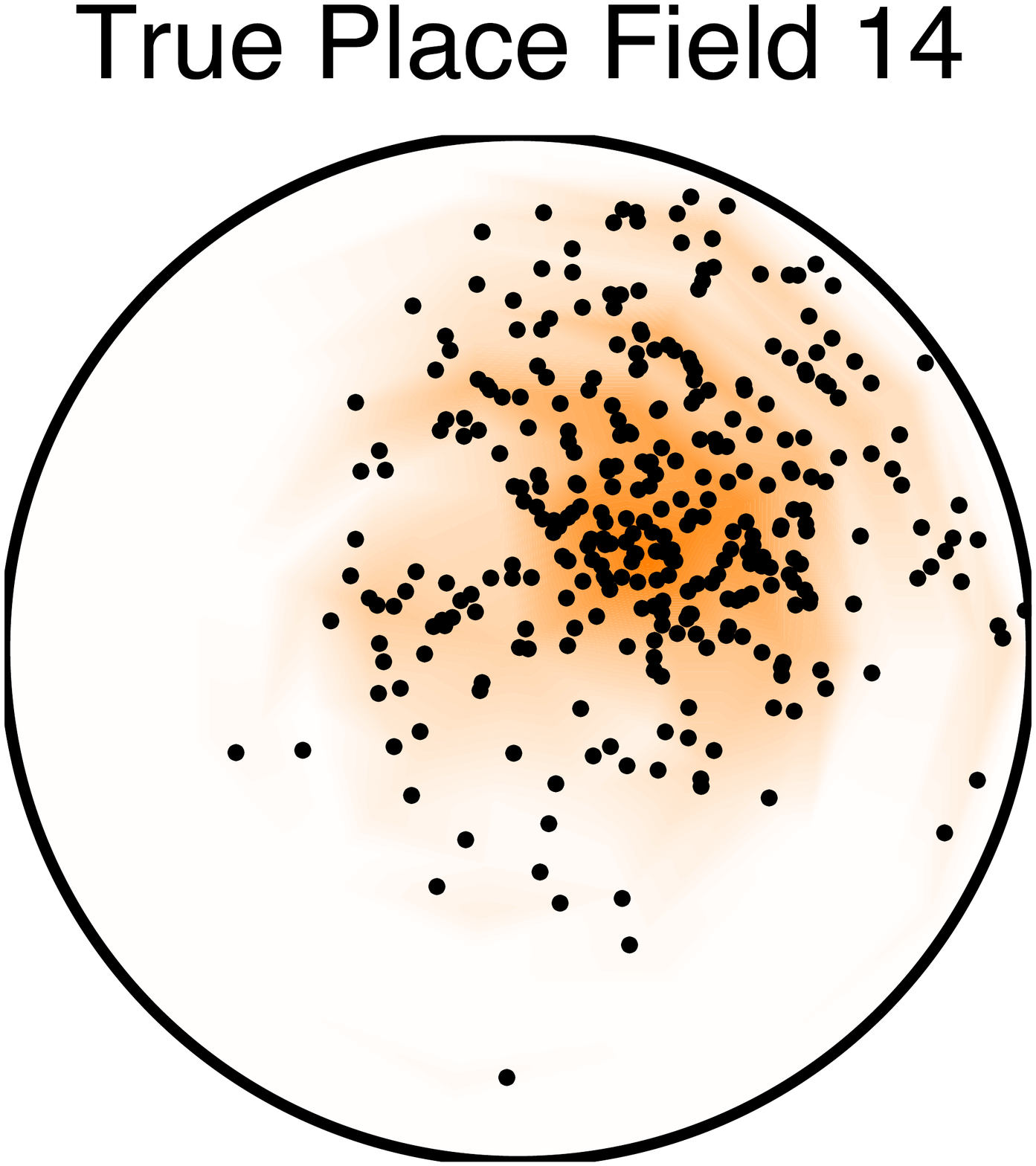}
\end{subfigure}
\\
\begin{subfigure}[t]{1.20in}
\includegraphics[width=\textwidth]{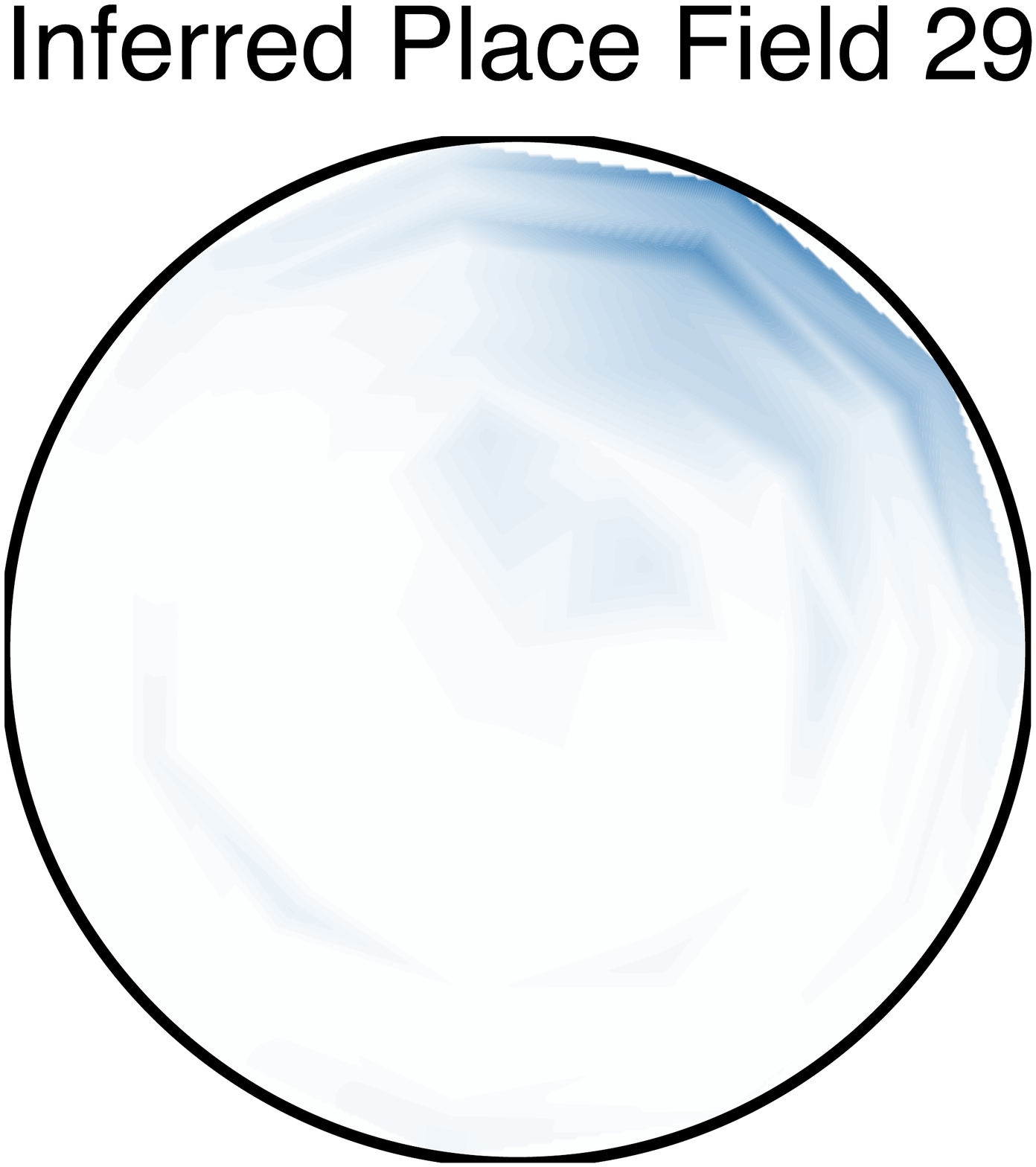}
\end{subfigure}
\hspace{-.1in}
~
\begin{subfigure}[t]{1.20in}
\includegraphics[width=\textwidth]{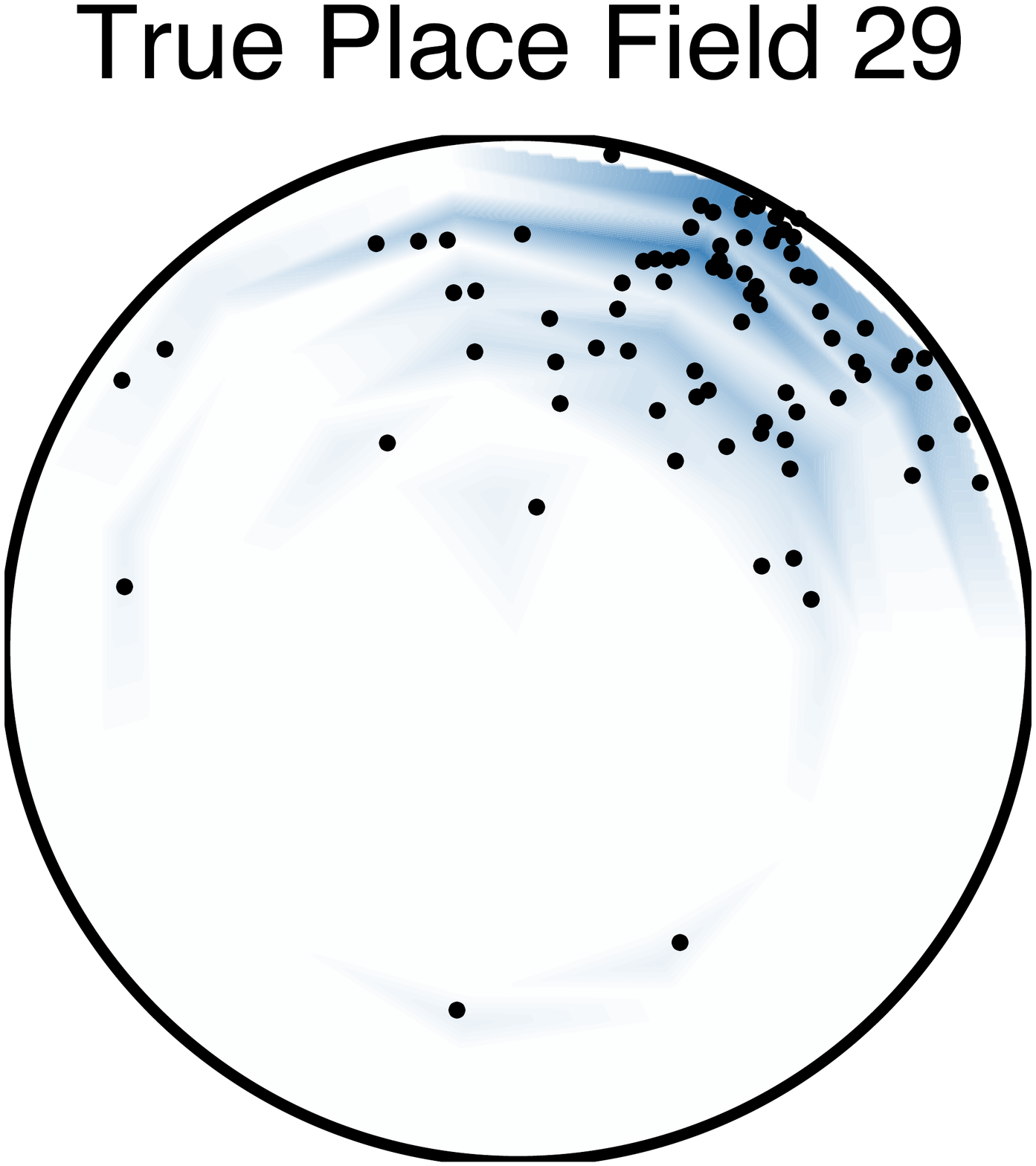}
\end{subfigure}
\hspace{-.1in}
~
\begin{subfigure}[t]{1.20in}
\includegraphics[width=\textwidth]{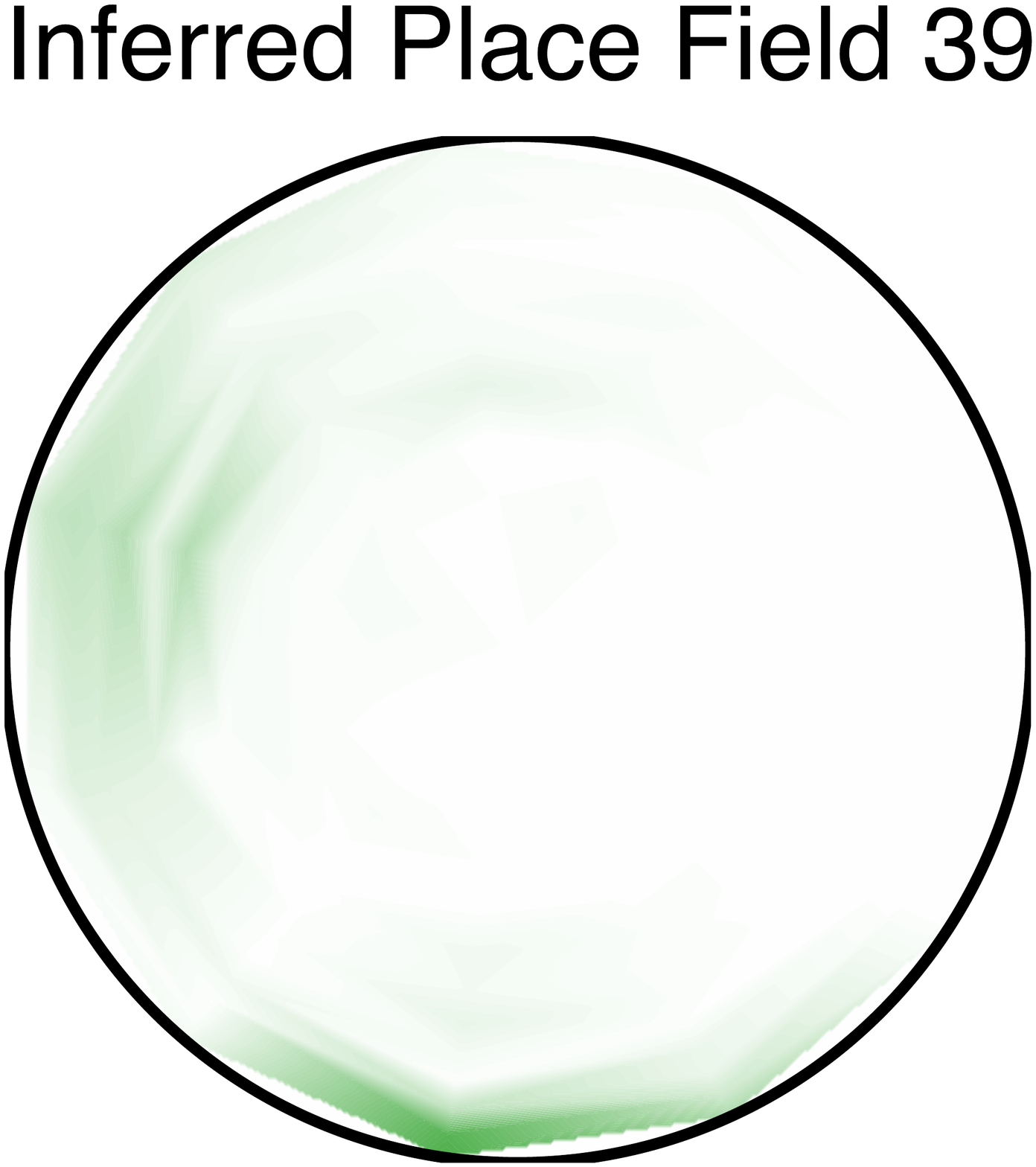}
\end{subfigure}
\hspace{-.1in}
~
\begin{subfigure}[t]{1.20in}
\includegraphics[width=\textwidth]{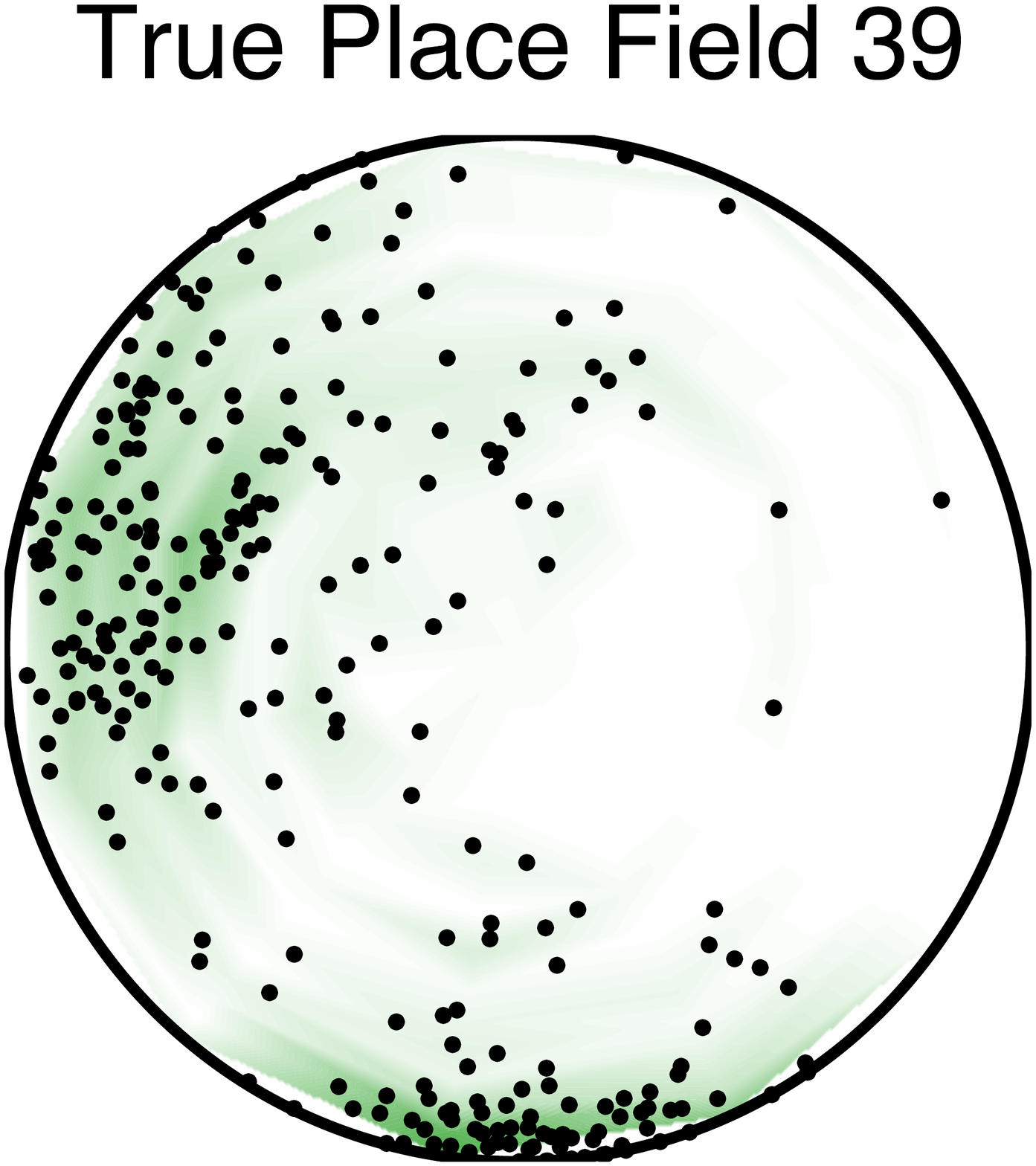}
\end{subfigure}
\caption{Pairs of inferred and true place fields for four randomly selected neurons. The inferred place field for cell~$c$ is a combination of location distributions for each state~$i$ weighted by the inferred firing rates~$\lambda_{c,i}$, whereas the true place field for cell~$c$ is a histogram of locations in which cell~$c$ fires. The black dots show the locations used for each histogram. We see that the inferred place fields closely match the true place fields. With adequate data, we expect a higher latent state dimensionality to yield higher spatial resolution in the inferred place fields.}
\label{fig8}
\end{figure}

As the parameter sample traces in Fig.~\ref{fig7} show, the Markov chains from HDP-HMM (MCMC+HMC) and HDP-HMM (MCMC+EB) converge in around 7500 iterations. After this point, the total number of states stabilizes to~${125.5\pm 2.0}$, but the number of states that account for 95\% of the time bins stabilizes to~${77.1\pm 1.8}$. 
The concentration parameters~$\alpha_0$ and~$\gamma$ converge within a similar number of iterations.
Finally, we show the  transition matrix~$\bP$ and firing rate matrix~$\bLambda$ obtained from the final sample.

\begin{table}
\centering
\caption{A comparison of HDP-HMM models and algorithms on the rat hippocampal data. Performance is measured in mean decoding error and predictive log likelihood on two minutes of held out test data (the best result is marked in bold font).} 
\begin{tabular}{lccc}
\hline
 &  \specialcell{HDP-HMM\\(MCMC+HMC)} & \specialcell{HDP-HMM\\(MCMC+EB)} & \specialcell{HDP-HMM\\(VB)}\\
\hline
Decoding error (cm) &
{$\mathbf{9.07 \pm 5.14}$}  & $9.50 \pm 5.73$ & $10.41 \pm 5.69$ \\
\specialcell{Predictive log likelihood (bits/spike)}  & 
{$\mathbf{0.591 \pm0.008}$} & ${0.530\pm0.010}$ & $0.581\pm0.005$ \\
\hline
\end{tabular}
\label{table2}
\end{table}

We perform a quantitative comparison between the three nonparametric models in terms of both decoding error and predictive log likelihood. For both tests, we train the models on the first $7.8$~minutes of data and test on the final two minutes of data for prediction. The results are summarized in Table~\ref{table2}. We find that the HDP-HMM (MCMC+HMC) again outperforms the competing models in both measures, though the differences in decoding performance are not statistical significant.

Looking into the inferred states, we can reconstruct the ``place fields'' of the hippocampal neurons, that is the distribution over locations of the rat given that a neuron fired. To do so, we combine the state-location maps of Fig.~\ref{fig6} (top right) with the firing rate of the neuron of cell~$c$ in those states and weight by the marginal probability of the latent state. Together, these give rise to the inferred neuron's place field, where, again, the position data was only used in reconstruction but not in the inference procedure. Four pairs of inferred and true place fields are shown in Fig.~\ref{fig8}. On the left is the inferred place field; on the right is the true place field computed using the locations of the rat when cell~$c$ fired shown by black dots.

\begin{figure}
\centering
\begin{subfigure}[t]{2.40in}
\includegraphics[width=\textwidth]{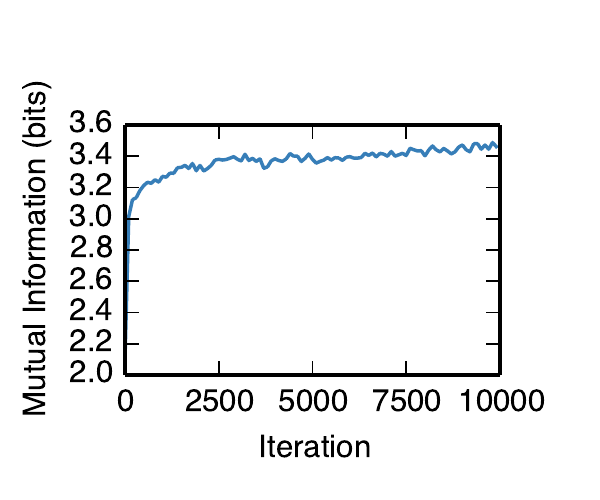}
\end{subfigure}
~
\begin{subfigure}[t]{2.40in}
\includegraphics[width=\textwidth]{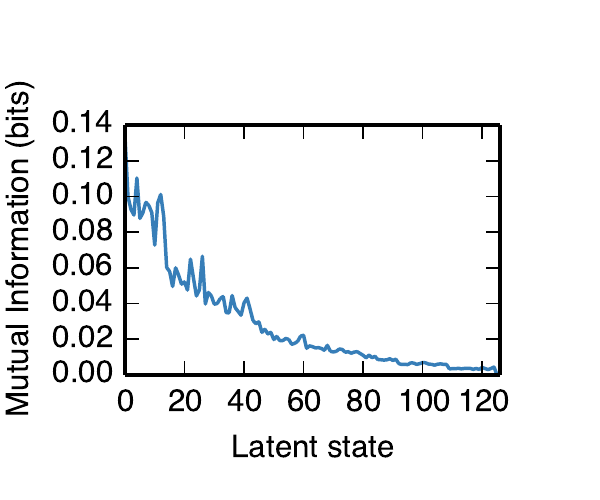}
\end{subfigure}
\caption{Mutual information of the inferred states and the rat's position. On the left, we show the mutual information of the state sequence and the rat's position (discretized into 121 equal-sized bins). On the right, we show how much information each state conveys about the location, ordered by the occupancy of the states. As expected, low-indexed states, i.e. those states which are most occupied, carry the most information. }
\label{fig9}
\end{figure}

In addition, we can evaluate the model in terms of the information latent states convey about the rat's position in the circular environment. To do so, we divide the environment into 121 bins of equal area and treat the rat's position as a discrete random variable. Likewise, we treat the latent state as a discrete random variable, and we compute the discrete mutual information between these two variables. The left panel of Fig.~\ref{fig9} shows how this mutual information increases with increasing MCMC iteration. As the model refines its estimates of the latent states, the latent state sequence carries greater information about position. We also investigate the information content of each individual state by constructing a binary random variable indicating whether or not the model is in state~$i$ and measuring its mutual information with the rat's position. The result is shown in the right panel of Fig.~\ref{fig9}, where the latent states are ordered in decreasing order of occupancy. As expected, states that are more frequently occupied  carry the most information about the rat's position.

\section{Extensions and Discussion}

\subsection{Hidden Semi-Markovian Models}
A striking feature of the inferred transition matrix in Fig.~\ref{fig7} is that the first 60 states, those which account for about~95\% of the time bins, exhibit strong self transitions. This is a common feature of time series and has been addressed by a number of augmented Markovian models. In particular, hidden semi-Markovian models (HSMMs) explicitly model the duration of time spent in each state separately from the rest of the transition matrix \citep{Johnson13}. Building this into the model allows the Dirichlet or HDP prior over state transition vectors to explain the rest of the transitions, which are often more similar. Alternatively, ``sticky'' HMMs and HDP-HMMs accomplish a similar effect \citep{Fox08}.  Our preliminary results (not shown) has already suggested that some improvement can be found in adopting this approach.

\subsection{Statistical and Computational Considerations}  In Bayesian estimation, we have seen a great advantage in nonparametric Bayesian formalism (i.e., HDP-HMM vs. HMM) regarding automatic model selection. This is especially important for sparse sample size or short recording in some neuroscience applications, where cross-validation on data is  infeasible.  

For any statistical estimation, we need to consider the ``bias vs. variance'' problem.  
In VB inference, there is a potential estimate bias due to bound optimization (since we optimize the lower bound of the marginal likelihood). In addition, because of the mean-field approximation, the parameter's variance tends to be underestimated. In MCMC inference, the estimate  is asymptotically unbiased, however, if the Markov chain mixes slowly, the estimate's variance can be inaccurate.

\subsection{Latent State Dimensionality}

In experimental data analysis, the number of identified  states from HDP-HMM depends on  the data. Given the same size of environment, different numbers of cells or different lengths of duration may yield different estimation results, since the nonparametric prior allocates states in accordance with the complexity of the data. We found that the weak priors over the concentration parameters have a minimal effect on the number of inferred states. The choice of firing rate hyperparameters does, however, have a strong effect. We also found that manually setting the firing rate hyperparameters to seemingly reasonable values could yield far fewer states. This is most likely because the unused states are assigned firing rates from the prior, and if the prior does not match the data, the sampled firing rates may not be able to explain the data and hence are unlikely to be chosen in favor of an existing state, whose firing rates are influenced by the data as well as the prior. However, it is also worth emphasizing that although the inferred state dimensionality may vary depending on different hyperparameters, the derived decoding error is insensitive to the exact number of state dimensionality.

\subsection{Robustness of the Population Firing Model}

A key assumption in our probabilistic model is the Poisson likelihood. Although this assumption may not be true in experimental data, our results have showed excellent performance.   
To further assess the robustness of HDP-HMM-Poisson model in experimental data analysis,
at every  temporal bin we  further add additional  homogeneous non-Poissonian noise  to the observed population spike counts by drawing from a NB distribution (with varying levels of mean 0.25-1.0 and variance 0.5-2.0), and repeat the decoding error analysis. We have found that, as a general trend,  the median decoding error gradually grows as increasing noise mean or variance; yet the decoding performance is still quite satisfactory (results not shown). 

\subsection{Use of Soft-labeled Spikes}

Thus far, we have assumed that all recorded ensemble spikes are sorted and clustered into single units. Nevertheless, it is known that spike sorting is complex, time-consuming and error-prone \citep{Wood08,Shalchyan14}.
On the one hand,  sorting error is inevitable when there is strong overlapping features (such as spike waveforms or principal components).
On the other hand,  traditional spike-sorting procedures often throw away considerable non-clusterable ``noisy'' spikes, which might contain informative tuning information \citep{Chen12b,Kloosterman14}. How to use these noisy spikes and maximize the information efficiency remains an open question. 
In other words, can we conduct the ensemble spike analysis using unsorted spikes?

Motivated from a sorting-free ensemble decoding analysis \citep{Chen12b,Kloosterman14}, we may use a soft-clustering method based on a Gaussian mixtures model (parameterized by an augmented   vector $\vect\xi=\{\ell_c,\vect\mu_c,\matr\Sigma_c\}_{c=1}^K$  that characterizes the weights, mean, and covariance parameters of the Gaussian mixtures). By clustering the spike waveform feature space, we assign each spike with a ``soft'' class label (about the unit identity) according to the posterior probability within the $K$-mixtures. In the feature space, the points close to (far away from) the $c$-th cluster center are associated with a probability assignment value close to (smaller than) 1 in the $c$-th class. Because of the soft membership of individual spikes, the spike count $y_{c,t}$ ($c = 1,\dots, K$) within a time interval can be a non-integer value. Consequently, we replace the variable $C$ with $K$ to indicate that the number of neurons is unknown, and rewrite the  log likelihood as follows   
\begin{eqnarray}                                   
\log p({\by}_{1:T}|\mathcal{S},\vect\theta,\vect\xi)= \sum_{t=1}^T \sum_{c=1}^K \log p(y_{c,t}|S_t,\vect\theta,\vect\xi)
\end{eqnarray}

In this case, the inference procedure consists of two steps: At the first stage, the $d$-dimensional spike waveform features are clustered 
using a ``constrained'' Gaussian mixture model \citep{Zou12}, which can be either finite or infinite. In the case of infinite Gaussian mixtures, we can also resort to the nonparametric Bayesian approach \citep{Rasmussen99,Gorur10,Wood08}. Upon completing the inference, each spike will be given a posterior probability of being assigned to each cluster. 
At the second stage, we sum the soft-labeled spikes  to obtain the probabilistic spike count $y_{c,t}$ for all $K$-clusters, and the  remaining 
nonparametric Bayesian (MCMC or VB) inference  procedure remains unchanged. A detailed investigation of this idea will be pursued in future work.

\section{Conclusion}

In this paper, we have explored the use of HDP-HMMs with Poisson likelihoods to analyze rat hippocampal ensemble spike data during spatial navigation. Compared to 
the parametric finite-state HMM, the HDP-HMM allows more flexibililty to model the experimental data (without relying on time-consuming cross-validation in model selection). We evaluate two nonparametric Bayesian inference algorithms for HDP-HMM,  one based on VB and the other based on MCMC. Furthermore, we consider two approaches for hyperparameter selection, an issue that is particularly important for the real-life application.  It is found that the MCMC algorithm with HMC updates for the hyperparameters is  robust and achieves the best performance in all simulated and experimental data. Our investigation shows a promising direction in applying nonparametric Bayesian methods for ensemble neuronal spike data analysis.

\subsection*{Acknowledgements}
We thank the members of the Wilson laboratory for  sharing the experimental data and valuable discussions. S.W.L. was supported by a National Defense Science and Engineering Graduate (NDSEG) Fellowship and by the Center for Brains, Minds and Machines (CBMM). M.J.J. was supported by the Harvard/MIT Joint Research Grants Program.
M.A.W. was supported by the NIH Grant R01-MH06197, TR01-GM10498,  ONR-MURI N00014-10-1-0936 grant.  This material is also based upon work supported by the Center for Brains, Minds and Machines (CBMM), funded by NSF STC award CCF-1231216.
Z.C. was supported by an NSF-CRCNS award (No. IIS-1307645) from the National Science Foundation.

\bibliographystyle{unsrtnat}
{\small \bibliography{arxiv}}

\end{document}